\documentclass{article}
\PassOptionsToPackage{table}{xcolor}
\usepackage[T1]{fontenc}
\usepackage{iclr2027_conference,times}
\usepackage{float}
\usepackage{placeins}
\usepackage{tabularx}
\usepackage{booktabs}
\usepackage{longtable}
\usepackage{adjustbox}
\usepackage{xcolor}
\usepackage{xspace}

\usepackage{amsmath,amsfonts,bm}

\def\eqref#1{equation~\ref{#1}}

\def\1{\bm{1}}

\DeclareMathAlphabet{\mathsfit}{\encodingdefault}{\sfdefault}{m}{sl}
\SetMathAlphabet{\mathsfit}{bold}{\encodingdefault}{\sfdefault}{bx}{n}

\usepackage{float}
\usepackage{pifont}
\usepackage{footnote}
\usepackage{enumitem}
\usepackage{bm}
\usepackage{arydshln}
\usepackage{booktabs}
\usepackage{multicol}
\usepackage{multirow}
\usepackage{color}
\usepackage{xcolor}
\usepackage{colortbl}
\usepackage{soul}
\usepackage{bbding}
\usepackage{makecell}
\usepackage{mathtools}
\usepackage{imakeidx}
\usepackage{float}
\usepackage{placeins}
\usepackage{amssymb}
\usepackage{graphicx}
\usepackage{amsmath}
\usepackage{threeparttable}
\definecolor{citecolor}{HTML}{0071BC}
\definecolor{linkcolor}{HTML}{ED1C24}
\usepackage[colorlinks,
  anchorcolor=red,
  citecolor=citecolor,
  linkcolor=linkcolor,
]{hyperref}
\makeindex
\usepackage{arydshln}
\usepackage{lipsum}
\usepackage[toc]{multitoc}
\usepackage[edges]{forest}
\usepackage[normalem]{ulem}

\usepackage{bbding}

\usepackage{algorithm}

\usepackage[toc,page,header]{appendix}

\definecolor{highlight}{rgb}{0, 255, 0}

\newcommand{\dataset}{{\fontfamily{ppl}\selectfont
CancerVerse}}

\newcolumntype{P}[1]{>{\centering\arraybackslash}p{#1}}
\newlength\savewidth

\usepackage{hyperref}
\usepackage{url}
\usepackage{pifont}

\usepackage{multirow}
\usepackage{amsmath}
\usepackage{algorithm}
\usepackage{algpseudocode}
\newcommand{\cropsize}{128}
\newcommand{\cropvol}{$\cropsize^3$}
\newcommand{\cropfull}{$\cropsize{\times}\cropsize{\times}\cropsize$}

\newcommand{\method}{DART\xspace}
\newcommand{\methodfullname}{Diffusion for Anatomical Reshaping with Text}
\renewcommand{\dataset}{DART-45K\xspace}
\newcommand{\segmodel}{DART-Seg\xspace}

\providecommand{\best}[1]{\textbf{\textcolor{blue}{#1}}}
\renewcommand{\best}[1]{\textbf{\textcolor{blue}{#1}}}
\providecommand{\second}[1]{\textcolor{blue}{#1}}

\renewcommand{\second}[1]{\textcolor{blue}{#1}}
\newcommand{\gain}[1]{\textcolor[HTML]{2E9B3F}{\textbf{#1}}}

\title{\method: Scaling a Shape Representation Across\\Anatomical Classes and Imaging Modalities}

\author{
  \begin{tabular}[h]{l} 
    \bfseries Koushik Swarna$^{1}$ \quad Pedro R. A. S. Bassi$^{1}$ \quad Lam Ngo$^{2,3}$ \quad Wenxuan Li$^{1}$ 
  \end{tabular} \\ 
  \begin{tabular}[h]{l} 
    \bfseries ~Jiancheng Yang$^{2,3}$ \quad Alan Yuille$^{1}$ \quad Zongwei Zhou$^{1,4}$\thanks{Correspondence to Zongwei Zhou (\href{mailto:zzhou82@jh.edu}{\textsc{zzhou82@jh.edu}}).} 
  \end{tabular} \\ [0.15cm]
  \begin{tabular}[h]{ll}
    \normalfont $^{1}$Johns Hopkins University & \normalfont $^{2}$ELLIS Institute Finland \\ 
    \normalfont $^{3}$Aalto University &
    \normalfont $^{4}$Johns Hopkins Medicine
  \end{tabular}
}

\iclrfinalcopy 
\begin{document}
\maketitle
\lhead{Preprint} 

\begin{abstract}
Current state-of-the-art segmentation models still produce anatomically implausible shapes, such as
broken vessels and fragmented glands, on the structures that are hardest to
segment. Learned shape representations capture what structures look like and
can help correct these errors, but most cover a fixed set of anatomical classes.
Extending this coverage across classes and imaging modalities is costly because
it often requires new voxel-level annotations. We propose \methodfullname{}
(\method), which learns a shared shape representation from segmentation
probabilities. It uses anatomical descriptions to correct these probabilities
without receiving medical images. Across $13$ public benchmarks, absolute Dice gains reach
$31\%$ for individual gallbladder cases in CT and $10\%$ for individual prostate
cases in MRI. Using existing segmentation predictions, we scale a single shape
representation from $50$ to $377$ anatomical classes without new voxel-level
annotations for correction training. The shared representation also corrects
four lesion classes absent from training and transfers from CT to MRI without
retraining. Corrected labels also improve
segmentation models, raising one model's Dice from $57.1\%$ to $67.5\%$.
The same pipeline supports annotations for $45{,}000$ CT scans. These results suggest that sharing
shape information within and across classes can reduce the need for new
voxel-level annotations when extending segmentation to more anatomical classes
and imaging modalities. The model is available at \href{https://huggingface.co/Koushik45048545309/DART-Seg}{https://huggingface.co/Koushik45/DART-Seg}.
\end{abstract}

\section{Introduction}\label{sec:introduction}

\begin{figure}[!t]
    \centering
    \includegraphics[width=1\textwidth]{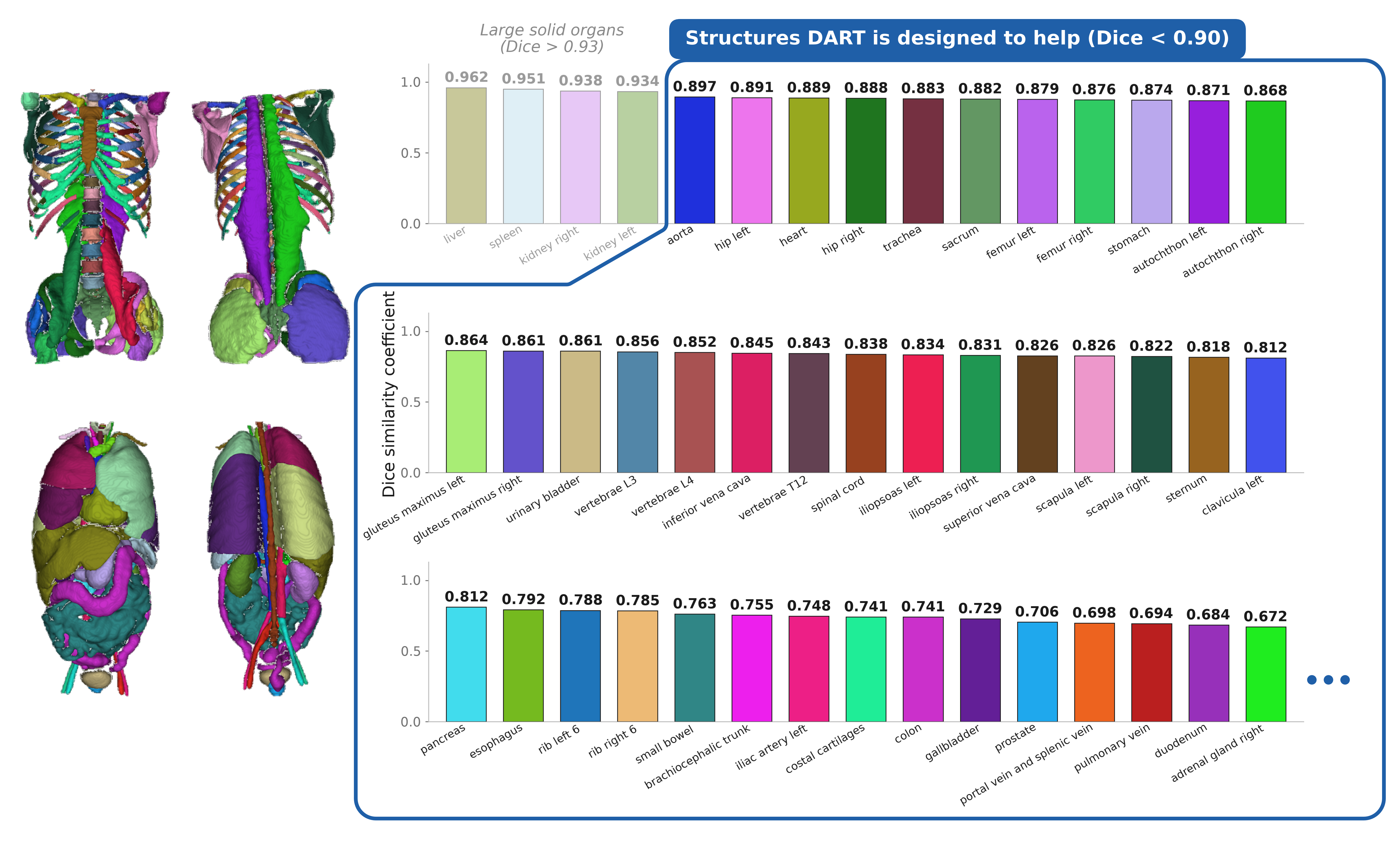}
    \caption{\textbf{Segmentation quality is uneven across anatomy.} Best
    reported Dice per class, from the Touchstone benchmark
    \citep{bassi2024touchstone} and TotalSegmentator
    \citep{wasserthal2023totalsegmentator}. Large solid organs such as the
    liver and spleen exceed $0.95$. Most other structures, including tubular
    structures, small glands and hollow organs, score lower. The outlined
    structures, all below $0.90$, are the ones \method is designed to help; the
    dots indicate that its vocabulary extends to $377$ classes.}
    \label{fig:teaser}
\end{figure}

Anatomical shape provides reusable knowledge for medical image segmentation.
Conventional multi-class segmentation models learn from voxel-level
annotations, and extending them to more classes or imaging modalities often
requires additional annotations \citep{li2024abdomenatlas, bassi2025learning}.
Even with this supervision, thin vessels,
small glands, and hollow organs remain difficult (Figure~\ref{fig:teaser}).
Their masks can contain broken vessels or fragmented glands, even when much of
the anatomy is correctly located. However, learned shape representations used
for segmentation correction usually cover one anatomy or a fixed class set
\citep{larrazabal2020postdae, painchaud2020cardiac}, limiting the reuse of
existing annotations. We therefore ask: can a shared shape representation make
these annotations useful across patients, anatomical classes, and imaging
modalities?

We hypothesize that shape similarities within and across anatomical classes
allow knowledge learned by existing segmentation models to support correction
beyond their original training examples. Within a class, recurring geometry
lets predictions for some patients guide correction in other patients. Across
classes, anatomical descriptions let one model share common shape features.
This reuses anatomical knowledge expressed in segmentation probabilities,
without directly relying on image appearance. This reuse offers a way to reduce the need for new
voxel-level annotations as anatomical and modality coverage expands.

To test this hypothesis, we propose \methodfullname{} (\method), a single 3D
diffusion model that learns from segmentation probabilities for $377$
anatomical classes. Given a predicted segmentation probability map and a
description of the target anatomy, it corrects the probabilities using the
learned shape representation (Figure~\ref{fig:pipeline}). This preserves
information that would be lost by first converting the prediction to a binary
mask. The description specifies the class, shape, and
location needed to guide correction. The medical image is not required.
The design supports three capabilities.
\begin{itemize}[leftmargin=1.2em,itemsep=1pt,topsep=2pt]
    \item \textbf{Scaling across anatomical classes.} Anatomical descriptions
    let one model share shape information across classes. We test whether this
    representation can correct lesion classes absent from training, without
    collecting training annotations for those classes.
    \item \textbf{Transfer across imaging modalities.} The shape representation
    is learned from segmentation probabilities without image intensities. We test
    whether a model trained on probabilities predicted from CT can correct
    predictions from MRI without MRI
    training annotations, with its weights and noise-level selection rule
    unchanged.
    \item \textbf{Compatibility with segmentation models.} Correction operates
    on segmentation probabilities from automatic segmentation models or MedSAM
    with bounding-box prompts. Corrected probabilities are converted to masks
    that can serve as training labels for segmentation models on additional
    scans.
\end{itemize}

Our evaluation includes $13$ public benchmarks, ten external to the training
data. We compare with Post-DAE \citep{larrazabal2020postdae}, SegRefiner
\citep{wang2023segrefiner}, a single-pass U-Net with \method's backbone, and traditional methods
including morphological closing and statistical shape models. In the
controlled $90$-class comparison, \method achieves the highest Dice on $77$
classes
(Section~\ref{sec:beats_classical}). The CT-trained correction model also improves MRI masks
and four lesion classes absent from its training data, without additional
training annotations or retraining (Section~\ref{sec:scale_modality}). By learning
from existing segmentation predictions, we expand the training vocabulary from
$50$ to $377$ anatomical classes without new voxel-level annotations for
correction training. Paraphrase and mismatch tests show that correction responds
to the anatomical content of descriptions (Section~\ref{sec:scale_class}). Corrected labels also
improve all fifteen segmentation model configurations we train, spanning
convolutional, residual, attention-gated and transformer architectures, by up to
$10.4$ Dice points (Section~\ref{sec:downstream}). Together, these results support reusing shape
information in segmentation models' predictions both to correct new cases
and to produce training labels.

In summary, we make four contributions.
\begin{itemize}[leftmargin=1.2em,itemsep=1pt,topsep=2pt]
    \item We introduce a shared anatomical shape representation learned by a
text-conditioned 3D diffusion model. It learns from and corrects segmentation
probabilities using anatomical descriptions without receiving medical images
(Figure~\ref{fig:pipeline}; conditioning in Eq.~\ref{eq:conditioning},
training in Eq.~\ref{eq:objective}, correction in Section~\ref{sec:refinement}).
\item We demonstrate scaling from $50$ to $377$ anatomical classes
(Section~\ref{sec:scale_class}, Table~\ref{tab:structure_scaling}) and
transfer from CT to MRI and to unseen lesion classes without target-specific
training annotations or retraining (Figure~\ref{fig:zeroshot}a,
Appendix~\ref{app:external}).
\item We evaluate segmentation correction on $13$ public benchmarks against
learned and traditional correction methods (Table~\ref{tab:headline},
Figure~\ref{fig:organ_results}, Table~\ref{tab:allstructures}), with inputs
from both automatic models (Figure~\ref{fig:organ_results},
Tables~\ref{tab:transfer_ts} and~\ref{tab:transfer_aa}) and prompted MedSAM
(Table~\ref{tab:headline}).
\item We show that corrected labels improve all fifteen segmentation model
configurations we train (Figure~\ref{fig:zeroshot}b,
Table~\ref{tab:longer}), and release \dataset, annotations for $45{,}000$ CT
scans (Appendix~\ref{app:downstream}).
\end{itemize}

\section{Related Work}\label{sec:related}

\textbf{Shape representations for segmentation.} Rule-based postprocessing, such as hole
filling, morphological closing and keeping the largest connected component, is
fast but applies the same rule to every structure
\citep{soille2004morphology, serra1983image, he2009fastccl}; organ-specific rule
sets narrow this gap but must be written by hand for each class
\citep{liu2025shapekit}. Statistical shape
models project a mask onto a linear space of training shapes
\citep{cootes1995asm, rogers2002robustasm}. Learned models are more flexible:
anatomically constrained models add a learned shape code to the training loss
\citep{oktay2018acnn}, Post-DAE maps an implausible mask to a plausible one with
a denoising autoencoder \citep{larrazabal2020postdae},
\citet{painchaud2020cardiac} map a cardiac mask to the nearest valid shape in a
latent space, and implicit shape models learn shape manifolds or templates end
to end \citep{raju2022deep, yang2022implicitatlas}. Neural Annotation Refinement
improves adrenal gland annotations \citep{yang2022near}, and template-guided
reconstruction shows that anatomical plausibility matters beyond voxel overlap
\citep{xie2025impulse}. Each of these models covers one anatomy, such as the
pancreas \citep{yao2022unsupervised}, or a fixed set of anatomical classes. SegRefiner adjusts predicted segmentation masks with a
discrete diffusion process but does not know which structure it is adjusting
\citep{wang2023segrefiner}, and in natural images CascadePSP and SegFix sharpen
mask boundaries using the image \citep{cheng2020cascadepsp, yuan2020segfix}.
\method is a single model that knows which class it is correcting, because a
description names it, and it needs no image. Quality-estimation methods score a
mask without reference annotations rather than correcting it
\citep{cosarinsky2026confic, galati2021efficient, wang2020qcgen, chen2026large}.

\textbf{Diffusion models in medical segmentation.} Denoising diffusion models
learn to reverse a gradual noising process \citep{ho2020ddpm}, and conditioning
steers one model toward different outputs \citep{dhariwal2021diffusion}. In
medical imaging they are mostly used to predict segmentation masks from images
\citep{wu2024medsegdiff, bieder2024diffusion, xing2023diffunet}, or to generate
images from text \citep{li2026text}. SDEdit edits an
input by partially noising and then denoising it \citep{meng2022sdedit}, and
diffusion models can reuse learned distributions in other inference problems
\citep{graikos2022plug}. \method applies this idea to segmentation
probabilities.

\textbf{Text-prompted and universal segmentation.} TotalSegmentator segments
more than one hundred structures from CT \citep{wasserthal2023totalsegmentator},
and nnU-Net remains a strong default \citep{isensee2021nnunet}. The CLIP-Driven
Universal Model, SegVol, SAT and VoxTell segment many structures from text
prompts \citep{liu2023clip, liu2024universal, du2024segvol, zhao2025sat,
rokuss2026voxtell}, and
SAM and MedSAM made promptable segmentation widely available
\citep{kirillov2023segment, ma2024medsam}. \method borrows the text interface
but does not segment images: it corrects the segmentation probabilities that
such models predict, and its corrected masks can serve as their training labels.

\section{Method}\label{sec:methods}

\begin{figure}[!t]
    \centering
    \includegraphics[width=\linewidth]{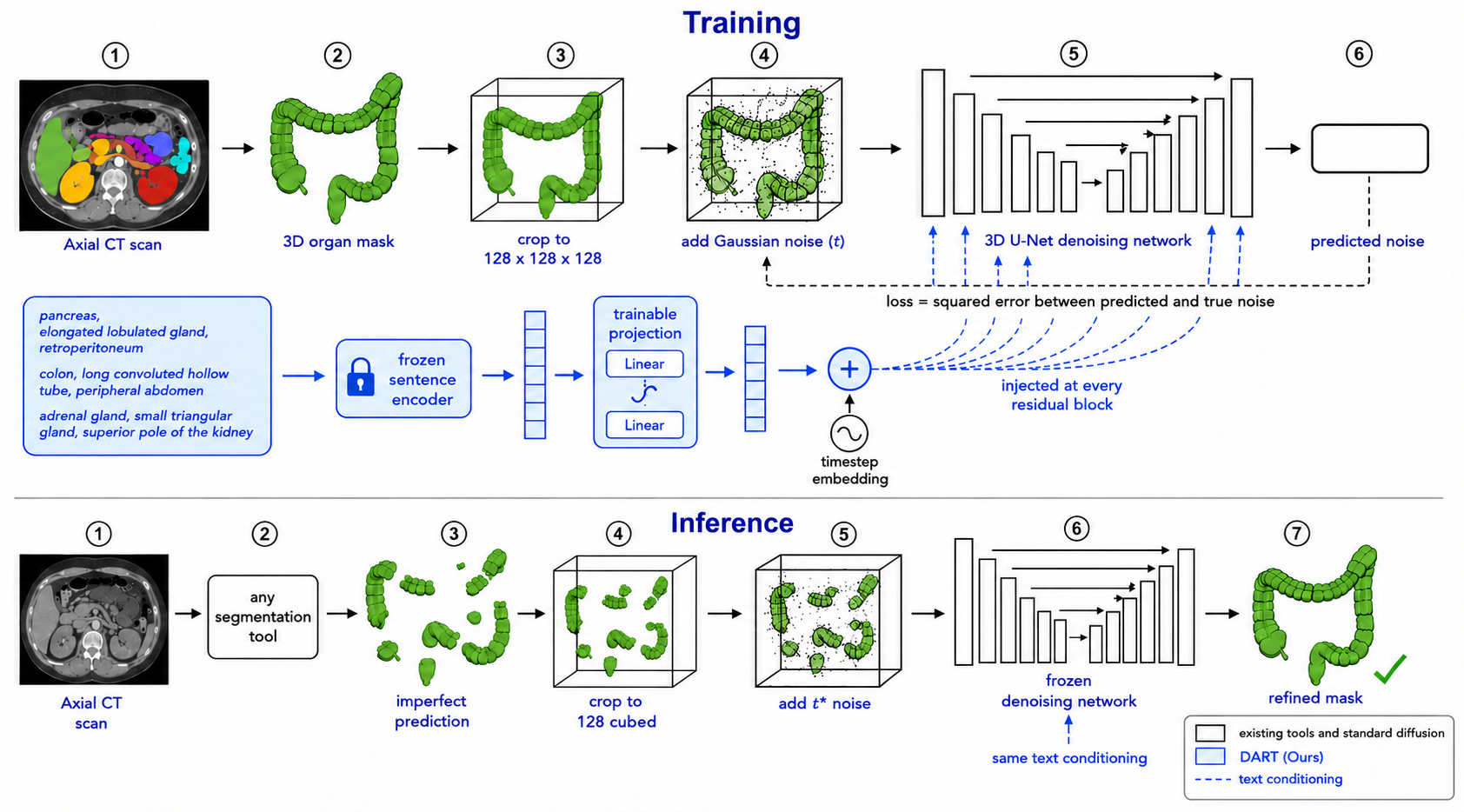}
    \caption{\textbf{Overview of \method.} \textit{Training} (top): a segmentation
    probability map predicted by a segmentation model is cropped, resized to
    \cropfull{}, and corrupted with Gaussian noise. A 3D U-Net learns to reverse
    the corruption. \textit{Conditioning} (middle,
    blue): an anatomical description is encoded by a frozen text encoder,
    projected by a trainable MLP, added to the timestep embedding, and injected
    into every residual block. \textit{Inference} (bottom): probabilities are
    thresholded at $0.5$ only to locate the crop. Their original values within the
    crop are resized, noised up to $t^*$, and denoised under the same description.
    The final result is thresholded to obtain a segmentation mask. Blue
    components are contributed by this work.}
    \label{fig:pipeline}
\end{figure}

\method learns from and corrects segmentation probabilities predicted by
segmentation models. For a class named by a description $s$, its input is a
probability map $p\in[0,1]^{H\times W\times D}$, where $p_v$ is the predicted
probability that voxel $v$ belongs to the class. It has three parts
(Figure~\ref{fig:pipeline}): a conditioning path that turns $s$ into a vector,
a 3D diffusion model that learns a shape representation from probability maps,
and a partial noising procedure that corrects $p$. Probability values preserve
information that a binary input would discard: a value of $0.4$, for example,
would otherwise become background after thresholding at $0.5$.

\subsection{Conditioning on anatomical descriptions}\label{sec:conditioning}

Each anatomical class is identified by a description of its name, shape, and
location, for example \textit{``colon, long convoluted haustrated hollow tube, framing the
peripheral abdomen from cecum to sigmoid''}. Descriptions are drafted by a large language model from
standard anatomy references and reviewed by the authors
(Appendix~\ref{app:descriptions}).
A frozen text encoder $E$, \texttt{all-MiniLM-L6-v2}
\citep{reimers2019sentencebert, wang2020minilm}, maps $s$ to a 384-dimensional
embedding, and a trainable projection $g$ maps that embedding to a
conditioning vector,
\begin{equation}
c = g(E(s)) \in \mathbb{R}^{d}, \qquad
g:\ \mathrm{Linear}(384,128) \rightarrow \mathrm{SiLU} \rightarrow \mathrm{Linear}(128,d),
\label{eq:conditioning}
\end{equation}
where $d$ is the width of the U-Net timestep embedding. The timestep embedding
$\tau_t$ and $c$ are summed, $h = \tau_t + c$, and $h$ is injected into every
residual block as a per-channel offset. We use a text encoder rather than
an integer embedding table for two reasons. An index table has no entry for a
class outside its training vocabulary, and its entries carry no information
about how structures relate. A text encoder can encode descriptions of classes absent from training and
represent relationships between descriptions. Because the crop removes absolute
position and the model sees no neighbouring structures, location words cannot
place a structure in the body; they act only through the embedding, where they
make the descriptions of nearby structures similar. Section~\ref{sec:scale_class}
compares the two schemes.

\subsection{Learning the shape representation}\label{sec:representation}

During training, the Clean sample $x_0$ is constructed from a segmentation
probability map predicted by an existing segmentation model. We threshold the
map at $0.5$ to locate the structure, add a fixed margin to its bounding box,
and crop and resize the original probability values to \cropvol. The binary
localization mask is not the diffusion input. Here, \emph{clean} means before
Gaussian corruption, not an error-free reference annotation.
Following \citet{ho2020ddpm}, the forward process adds
Gaussian noise, $q(x_t \mid x_0) = \mathcal{N}\big(x_t;\
\sqrt{\bar\alpha_t}\,x_0,\ (1-\bar\alpha_t)\mathbf{I}\big)$, and the model
$\epsilon_\theta$ is trained to predict that noise under the conditioning vector $c$:
\begin{equation}
\mathcal{L} = \mathbb{E}_{x_0,\epsilon,t}\!\left[\left\lVert \epsilon - \epsilon_\theta(x_t,t,c)\right\rVert_2^2\right],
\qquad \epsilon \sim \mathcal{N}(0,\mathbf{I}).
\label{eq:objective}
\end{equation}
We use $T = 1000$ steps with a linear noise schedule
(Appendix~\ref{app:implementation}). The model parameters encode a shared
shape representation, expressed through
$p_\theta(x_0 \mid c)$, the conditional distribution of segmentation probability
maps. Conditioning on $c$ selects the anatomy specified by the description.
The diffusion model receives predicted probabilities, not medical images or
binary reference annotations. Its learned representation can reflect both
anatomical shape and errors in the source predictions. We use generic Gaussian
corruption rather than explicitly simulated segmentation errors.
Clean probability values lie in $[0,1]$; Gaussian-corrupted states $x_t$ need
not remain in this interval.

\subsection{Correction by partial noising}\label{sec:refinement}

At inference, we form a localization mask $m_v=\mathbf{1}[p_v>0.5]$ and
compute its bounding box with the same fixed margin. Thresholding determines
only this rough location. We Crop and resize the original probability map $p$
to obtain $x_0^{\mathrm{in}}$, retaining values below $0.5$ inside the crop.
If the localization mask is empty, we return an empty segmentation mask.
\method also receives the anatomical description and noise level $t^*$; it
receives no medical image, reference annotation, or bounding-box prompt.
Following SDEdit \citep{meng2022sdedit}, we add noise to the cropped probability
map up to an intermediate timestep $t^*$,
\begin{equation}
x_{t^*} = \sqrt{\bar\alpha_{t^*}}\,x_0^{\mathrm{in}} + \sqrt{1-\bar\alpha_{t^*}}\,\epsilon,
\qquad \epsilon \sim \mathcal{N}(0,\mathbf{I}),
\label{eq:forward}
\end{equation}
and run the reverse process from $t^*$ to $0$ under $c$. The denoised estimate
is resampled to the native geometry and thresholded at $0.5$ to obtain the
corrected segmentation mask $\hat m$
(Algorithm~\ref{alg:refine}, Appendix~\ref{app:algorithm}). A small $t^*$
allows limited changes, while a larger $t^*$ allows larger shape changes at the
cost of fidelity to the input.

\textbf{Choosing $t^*$.} The noise level $t^*$ is the only hyperparameter
specific to \method. It is selected once per class on a validation split
from $\{50, 150, \dots, 650\}$ and then held fixed for every test case and
every experiment in this paper, including the transfer experiments
(Figure~\ref{fig:noise}). The selected values follow anatomy: compact solid
organs take the least noise, glands and hollow organs more, and vessels and
muscles the most. For a class absent from the training vocabulary, such as
a lesion, $t^*$ is copied from the structure in which it occurs, for example
the hippocampus for a hippocampal tumor. No labels from a test task are used to
choose $t^*$. The same values work unchanged on segmentation probabilities from every
segmentation model we test (Appendix~\ref{app:transfer}), which suggests that
$t^*$ reflects the shape of a structure rather than the errors of one
segmentation model.

\section{Experimental Setup}\label{sec:experiments}

\textbf{Training data.} The diffusion model is trained on segmentation
probability maps predicted for $3{,}066$ CT, cone-beam CT (CBCT) and CT
angiography volumes from eight public datasets (Appendix~\ref{app:data}). Their
voxel-level annotations define the class inventory: merging labels that refer to
the same anatomy turns $420$ annotated label classes into a vocabulary of $377$
anatomical classes, spanning organs, tissues, body regions and composite
classes. No MRI volume is used, and the four lesion classes in these sources are
held out.

\textbf{Benchmarks.} We evaluate on $13$ public benchmarks. Three use held-out
subjects from training datasets: TotalSegmentator ($90$ classes), ToothFairy
($77$) \citep{bolelli2024toothfairy} and DAP-Atlas ($145$)
\citep{jaus2023dapatlas}. Ten are external and contribute nothing to training:
FLARE22 \citep{ma2024flare22}, KiTS23 \citep{heller2023kits21} and the spleen
and pancreas tasks of the Medical Segmentation Decathlon (MSD)
\citep{antonelli2022msd} in CT; the MSD heart, hippocampus and prostate tasks in
MRI; and the CT, T1 and T2 scans of CHAOS \citep{kavur2021chaos}. Held-out WORD
and AMOS22 cases, the PanSegNet MRI pancreas and seven transfer tasks complement
them (Appendix~\ref{app:protocol}).

\textbf{Input predictions.} \method receives the probability map predicted by a
segmentation model, never the medical image. On CT, three automatic segmentation
models predict without bounding-box prompts: TotalSegmentator
\citep{wasserthal2023totalsegmentator}, AtlasNet \citep{bassi2025radgpt,
li2024abdomenatlas} and WholeBodySeg \citep{cardoso2022monai}. MedSAM
\citep{ma2024medsam}, prompted slice by slice with bounding boxes from the
reference annotation, places every structure in the right location and so
isolates shape. On MRI we use MedSAM and TotalSegmentator MRI
\citep{dantonoli2025totalsegmentatormri}.

\textbf{Metrics.} Dice, IoU, clDice (centerline overlap, which reflects
connectivity) and NSD (boundary agreement within $2$\,mm) are measured against
the reference annotation on a $0$--$100$ scale, and their changes are absolute
points; HD95 is the $95$th-percentile Hausdorff distance in mm. ``Up to''
denotes the largest per-case gain unless stated otherwise
(Appendix~\ref{app:protocol}).

\textbf{Baselines.} Every correction method starts from the same upstream
predictions (\method from the probability map, the others from its thresholded
mask) and is tuned per class on the validation split, as $t^*$ is for \method.
On the TotalSegmentator benchmark we compare the eleven baselines of
Table~\ref{tab:headline}: four rules, two statistical shape models, two methods
that read the image, and three learned correction methods trained on predictions
paired with reference annotations: Post-DAE, SegRefiner and a single-pass U-Net
with \method's backbone but no diffusion. The external benchmarks add
further methods and the identity control (Appendix~\ref{app:baselines}).

\section{Results}\label{sec:results}

\subsection{\method corrects predictions from automatic and prompted models}\label{sec:beats_classical}

\begin{table}[!t]
\centering
\scriptsize
\setlength{\tabcolsep}{3pt}
\renewcommand{\arraystretch}{1.05}
\caption{\textbf{\method gives the largest gains on thin, small and elongated
structures.} Dice on the six TotalSegmentator classes with the largest \method
gains. Every method starts from the MedSAM predictions in the first row. $\pm$: standard deviation of the
per-case paired gain over the input; $\Delta$: \method's mean gain in Dice
points. Best per column in \best{bold blue}, second best in \second{blue};
$^\dagger$ reads image intensity; `---': no result. All $90$ classes are in
Table~\ref{tab:allstructures}.}
\label{tab:headline}
\begin{adjustbox}{max width=\linewidth}
\begin{tabular}{@{}ll*{6}{c}@{}}
\toprule
Type & Method & Autoch. L & Autoch. R & Spinal cord & Adrenal R & Glut. min. L & Esophagus \\
\midrule
--- & MedSAM input~\citep{ma2024medsam} & 79.1 & 79.8 & 77.5 & 65.6 & 63.8 & 78.5 \\
\midrule
\multirow{3}{*}{Learned} & Single-pass U-Net~\citep{ronneberger2015unet} & 78.8$\pm$0.4 & 79.5$\pm$0.4 & 76.9$\pm$1.0 & 65.6$\pm$0.0 & 63.7$\pm$0.1 & 78.0$\pm$0.7 \\
 & SegRefiner$^\dagger$~\citep{wang2023segrefiner} & 75.1$\pm$6.2 & 74.9$\pm$5.5 & 73.9$\pm$3.8 & 64.6$\pm$6.3 & 64.2$\pm$3.3 & 73.2$\pm$7.3 \\
 & Post-DAE~\citep{larrazabal2020postdae} & 62.4$\pm$14.4 & 62.7$\pm$14.8 & 61.8$\pm$12.4 & 39.0$\pm$8.5 & 51.4$\pm$5.8 & 58.0$\pm$18.4 \\
\midrule
\multirow{8}{*}{Traditional} & Hole filling~\citep{soille2004morphology} & 79.1$\pm$0.0 & 79.8$\pm$0.0 & 77.5$\pm$0.0 & 65.6$\pm$0.0 & 63.8$\pm$0.0 & 78.5$\pm$0.0 \\
 & Morphological closing~\citep{serra1983image} & 81.8$\pm$1.6 & \second{82.2$\pm$1.7} & \second{80.0$\pm$1.9} & \second{66.6$\pm$4.2} & \second{66.5$\pm$1.5} & \second{80.2$\pm$6.4} \\
 & Convex hull~\citep{barber1996quickhull} & \second{81.9$\pm$1.2} & 82.0$\pm$1.0 & 78.5$\pm$1.0 & \second{66.6$\pm$1.8} & 64.6$\pm$2.1 & \second{80.2$\pm$0.8} \\
 & Largest component~\citep{he2009fastccl} & 79.1$\pm$0.0 & 79.7$\pm$0.2 & 72.6$\pm$8.9 & 64.4$\pm$5.0 & 63.8$\pm$0.1 & 74.1$\pm$9.2 \\
 & Geodesic contour$^\dagger$~\citep{caselles1997geodesic} & 74.6$\pm$7.7 & 74.1$\pm$7.2 & 63.6$\pm$20.1 & 1.4$\pm$14.9 & 19.9$\pm$14.0 & 34.3$\pm$16.7 \\
 & GMM graph-cut$^\dagger$~\citep{rother2004grabcut} & 66.8$\pm$18.4 & 66.2$\pm$17.6 & 77.4$\pm$18.8 & 57.9$\pm$10.9 & 64.0$\pm$3.4 & 72.2$\pm$10.8 \\
 & ASM~\citep{cootes1995asm} & 21.3$\pm$28.3 & 18.7$\pm$27.9 & 8.2$\pm$21.1 & --- & 11.1$\pm$19.9 & 6.6$\pm$21.7 \\
 & Robust ASM~\citep{rogers2002robustasm} & 20.7$\pm$25.1 & 18.3$\pm$24.4 & 6.3$\pm$18.2 & --- & 11.1$\pm$19.9 & 6.6$\pm$21.7 \\
\midrule
\multirow{2}{*}{Ours} & \textbf{\method} & \best{85.7$\pm$2.7} & \best{85.4$\pm$2.9} & \best{83.1$\pm$4.4} & \best{70.7$\pm$3.4} & \best{68.8$\pm$2.0} & \best{83.3$\pm$3.3} \\
 & \quad $\Delta$ over MedSAM input & \gain{+6.6} & \gain{+5.6} & \gain{+5.6} & \gain{+5.1} & \gain{+5.0} & \gain{+4.8} \\
\bottomrule
\end{tabular}
\end{adjustbox}
\end{table}

\textbf{Best on $77$ of $90$ classes in the controlled comparison.} With every
correction method starting from the same MedSAM predictions on held-out
TotalSegmentator cases, \method raises mean Dice over $90$ classes from $73.6$ to
$75.6$, is best on $77$ and gains significantly ($5\%$ false discovery rate) on
$84$ (Table~\ref{tab:allstructures}, Appendix~\ref{app:negative}). It is the
only method whose class mean never falls below the input; where a baseline
leads, the margin is at most $4.1$ points. On the six classes with the largest
gains, all thin, small or elongated structures (Table~\ref{tab:headline}),
\method adds $4.8$ to $6.6$ Dice points and leads the strongest baselines,
morphological closing and the convex hull, by $2.3$ to $4.1$ points. The learned
baselines do not help: SegRefiner and Post-DAE lower Dice on $87$ and $90$
classes, and the single-pass U-Net stays within $1.6$ points of its input. Image-based and shape-model baselines can fail badly, for example the
geodesic contour on the right adrenal gland ($1.4$).

\begin{figure}[!t]
    \centering
    \includegraphics[width=\linewidth]{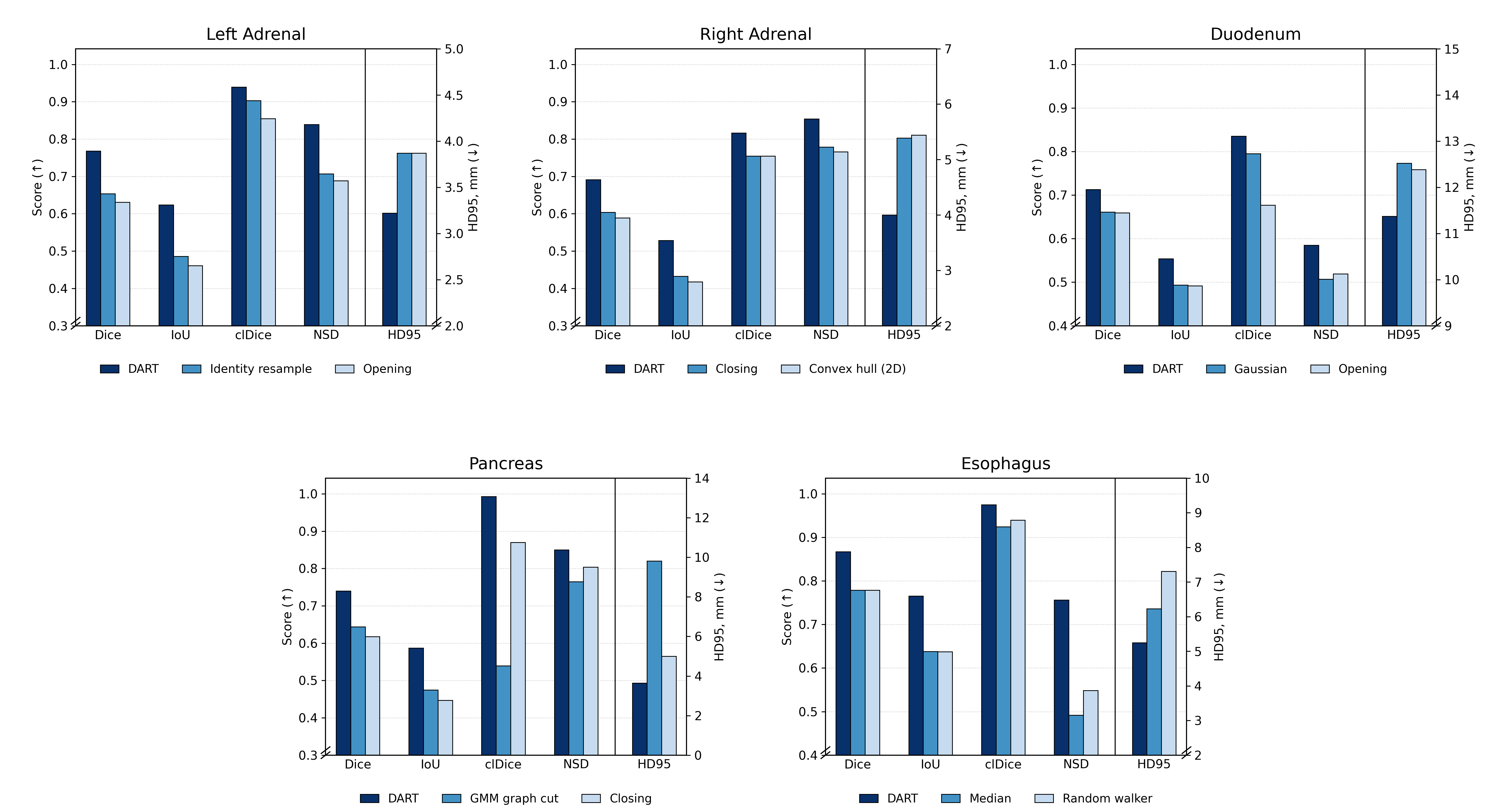}
    \caption{\textbf{On five hard structures, \method outperforms the two
    strongest baselines on every metric.} Predictions from TotalSegmentator,
    AtlasNet and WholeBodySeg on the external benchmarks, averaged over the
    three models and over the benchmarks that annotate each structure. Each
    panel compares \method with the two baselines with the highest mean Dice for
    that structure, named in its legend. Dice, IoU, clDice and NSD (within
    $2$\,mm) use the left axis (higher is better); HD95 in mm uses the right axis
    (lower is better).}
    \label{fig:organ_results}
\end{figure}

\textbf{Up to $31$ Dice points with automatic segmentation models.} With every
$t^*$ fixed in advance, the frozen model corrects predictions from
TotalSegmentator, AtlasNet and WholeBodySeg, which receive no bounding-box
prompts. On the external CT benchmarks, \method raises Dice by up to $31$ points
per case (gallbladder, FLARE22, AtlasNet input), and on five hard structures it
outperforms the two strongest baselines on every metric: Dice, IoU, clDice, NSD
and HD95 (Figure~\ref{fig:organ_results}). Mean Dice rises on every external
benchmark (Appendix~\ref{app:external}). On held-out WORD and AMOS22 cases it
rises by $2.47$ and $1.04$ points on TotalSegmentator and AtlasNet predictions,
and on held-out TotalSegmentator cases by $3.77$ points on WholeBodySeg
predictions of $104$ classes (Appendix~\ref{app:transfer}).

\textbf{Abnormal kidneys keep their tumors.} On the abnormal kidneys of KiTS23,
whose reference kidney includes the tumor, \method raises mean Dice on
TotalSegmentator predictions from $91.8$ to $95.1$ without erasing the tumor to
make the kidney look typical; giving the kidney another structure's description
lowers Dice by $4.9$ points (Appendix~\ref{app:external}).

\textbf{The gain comes from the diffusion model.} The identity control, which
skips diffusion, adds at most $0.24$ Dice points (Appendix~\ref{app:identity}),
whereas sampling from pure noise in the same crop, which discards the input,
scores $34.1$ Dice against an input of $77.4$ and \method's $78.9$ on about $50$
MSD heart and FLARE22 cases (Appendix~\ref{app:negative}). \method thus edits
its input rather than generating a class template. A single-pass U-Net with the
same backbone but no diffusion learns little more than the identity
(Appendix~\ref{app:negative}). On held-out WORD cases, \method also halves the connected components per mask, from $3.33$ to $1.67$
(Appendix~\ref{app:datasetlevel}).

\subsection{A shape representation learned from CT transfers to MRI}\label{sec:scale_modality}

\begin{figure}[!t]
    \centering
    \includegraphics[width=1\linewidth]{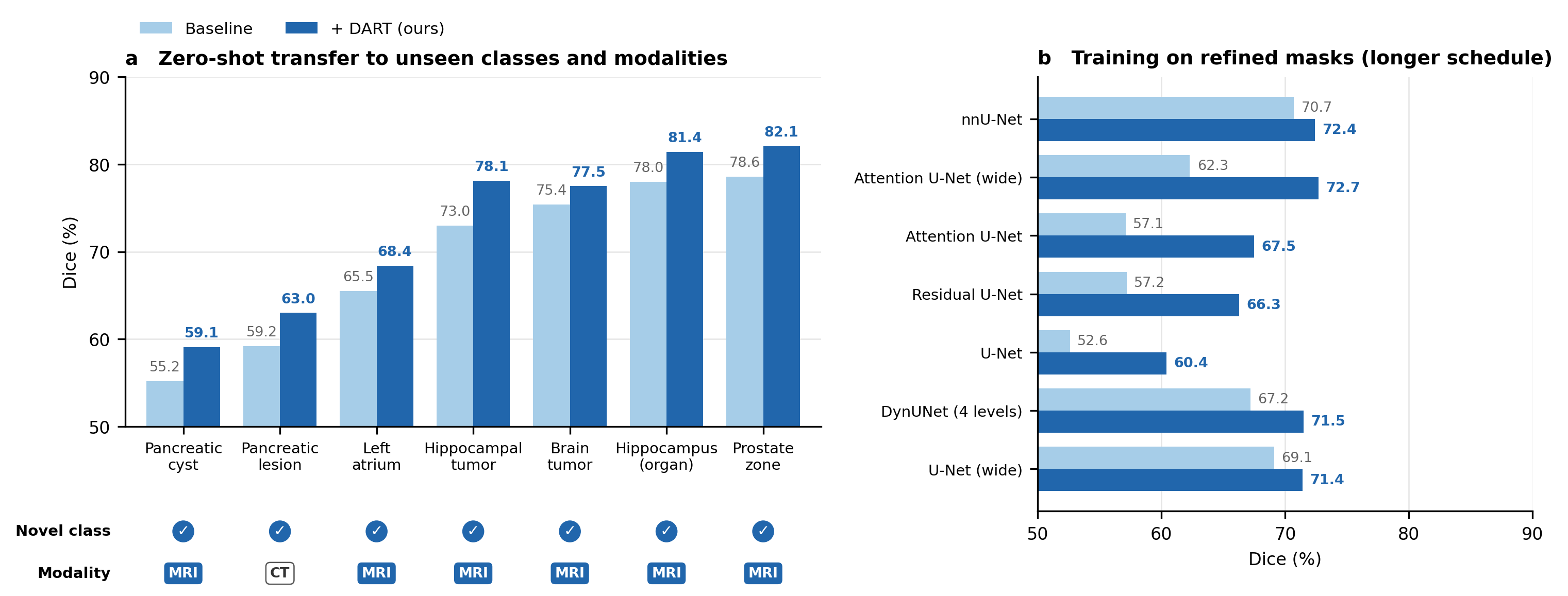}
    \caption{\textbf{One frozen model transfers to new tasks, and its corrected
    labels train better models.} \textbf{(a)} Dice before and after \method on
    the seven transfer tasks with MedSAM input, six in MRI and one in CT
    (pancreatic lesion); four are lesion classes, and none of the seven appears
    in training. \textbf{(b)} Seven of the fifteen configurations of
    Table~\ref{tab:longer}, trained on original and on corrected labels with
    every other factor fixed; nnU-Net is the five-level DynUNet.}
    \label{fig:zeroshot}
\end{figure}

\textbf{Up to $10$ Dice and $28$ NSD points on MRI.} Because \method never
receives image intensities, a model trained only on CT predictions can correct
MRI predictions; no MRI volume or lesion class appears in training, and neither
the weights nor any $t^*$ changes. On TotalSegmentator MRI predictions, mean Dice
rises on all four MRI-only datasets of Appendix~\ref{app:external}: the
PanSegNet MRI pancreas \citep{zhang2025pansegnet} from $74.1$ to $82.3$, the left
atrium from $80.2$ to $84.8$, the prostate from $87.6$ to $91.9$ and the
hippocampus from $86.0$ to $89.8$, with per-case prostate gains reaching $10$
Dice and $28$ NSD points. With MedSAM input, \method also improves all six MRI
tasks of Figure~\ref{fig:zeroshot}a, including pancreatic cysts from Cyst-X
\citep{pan2025cystx} in all seven scanner and sequence groups.

\textbf{Unseen lesion classes.} Four lesion classes in
Figure~\ref{fig:zeroshot}a never appear in training, three in MRI and one in CT,
the pancreatic lesions of PanTS \citep{li2025pants}. \method raises mean Dice on
all four, by $2.1$ to $5.1$ points, most on hippocampal tumor. We do not isolate
why. Early tumors share appearance across organs in CT
\citep{chen2024towards}, and a generic description such as ``rounded mass''
would separate knowledge of the named structure from generic smoothing.

\subsection{Corrected labels train better segmentation models}\label{sec:downstream}

\textbf{All fifteen configurations improve.} We train each segmentation model
configuration on masks predicted by the three automatic segmentation models,
TotalSegmentator, AtlasNet and WholeBodySeg (original labels), and on the masks
\method produces from the same predictions (corrected labels), with everything
else identical (Appendix~\ref{app:architectures}). Neither set of labels uses a
bounding-box prompt or a reference annotation. Corrected labels improve all
fifteen configurations (Table~\ref{tab:longer}, Figure~\ref{fig:zeroshot}b),
from nnU-Net to UNETR, so the gain comes from the labels rather than from one
design. Among the eleven configurations that train well, the gain reaches $10.4$
points (Attention U-Net, from $57.1$ to $67.5$), with a median of $3.9$.

\textbf{$45{,}000$ annotated CT scans.} We apply the same pipeline to $45{,}000$
public CT scans without annotations: published segmentation models predict each
structure, \method corrects every prediction, and \segmodel, a text-promptable
segmentation model trained on the corrected masks, produces the released
annotations, \dataset. In a reader audit of $10{,}000$ of these scans, $98.5\%$
passed (Appendix~\ref{app:downstream}).

\subsection{One shape representation scales across classes and training examples}\label{sec:scale_class}

\begin{figure}[!t]
    \centering
    \includegraphics[width=0.93\linewidth]{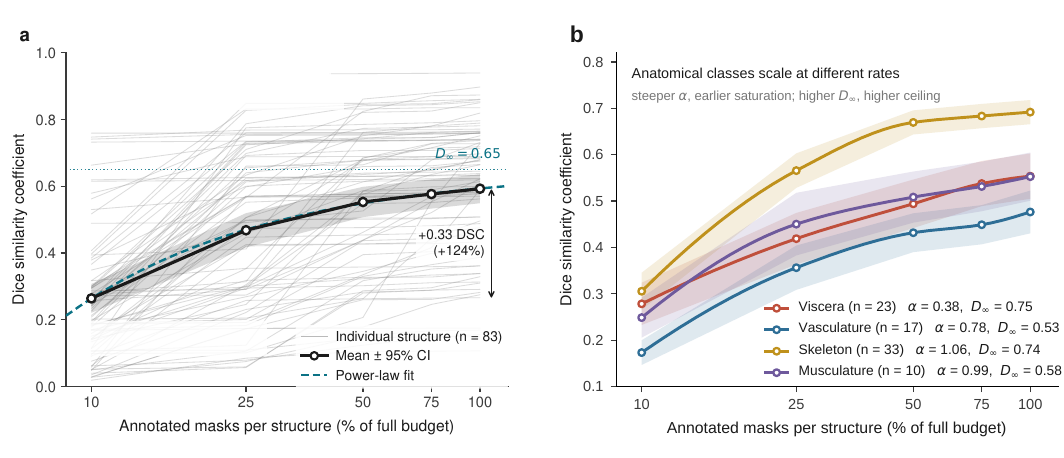}
    \caption{\textbf{The model keeps improving with more training examples.}
    \textbf{(a)} Dice after \method (on a $0$--$1$ scale) for $83$ anatomical
    classes when the model is trained on $10\%$ to $100\%$ of the available
    training examples: individual classes (thin lines), their mean with a $95\%$
    confidence interval (heavy line) and a saturating power-law fit to the mean
    (dashed; Appendix~\ref{app:scalingfit}). The mean curve passes the mean input
    Dice of $0.53$ at about $40\%$ of the training examples. \textbf{(b)} The same
    runs grouped by anatomy.}
    \label{fig:scaling}
\end{figure}

\textbf{Expanding the vocabulary preserves mean Dice.} A model trained on all
$377$ classes scores $74.20$ mean Dice on the original $50$ classes, against
$73.85$ for a model trained on those $50$ alone (Table~\ref{tab:structure_scaling}):
one model absorbs $327$ additional classes without lowering mean Dice on the
original ones.

\textbf{Descriptions outperform class indices, and \method follows them.}
Replacing the description with a learned class embedding, everything else fixed,
cuts the gain over the input from $5.49$ to $0.39$ Dice points
(Table~\ref{tab:indexing_comparison}), partly because an index cannot name
classes outside its training vocabulary; text embeddings likewise outperform
one-hot labels in multi-class segmentation \citep{liu2023clip, liu2024universal}. Rewording a description moves Dice by
at most $0.13$ points on any class, whereas another structure's description
lowers Dice on every class tested, by $2.07$ points on average
(Appendix~\ref{app:prompts}).

\textbf{More training examples make a better model.} Retrained on nested subsets
with $10\%$ to $100\%$ of each structure's training examples, the model's mean
Dice over $83$ classes rises from $26$ to $59$ and is still rising at the full
budget (Figure~\ref{fig:scaling}). It improves its input (mean $53$) from about
$40\%$ of the training examples onward; skeletal structures saturate earliest,
and thin, branching vessels reach the lowest ceiling (Appendix~\ref{app:scalingfit}).

\section{Discussion}\label{sec:discussion}

\textbf{Where \method helps most.} The gain follows the room the input leaves.
With automatic segmentation models, it falls as the input improves (Spearman
$\rho=-0.89$ and $-0.68$; Appendix~\ref{app:transfer}), and in
Appendix~\ref{app:external} it ranges from $8.2$ points on an input of $74.1$ to
$0.4$ on an input above $94$. In the controlled comparison, the largest gains fall on elongated
structures and small glands, whereas the portal vein, pulmonary vessels and colon
gain under $1$ point, consistent with the low ceiling of vessels in
Figure~\ref{fig:scaling}b.

\textbf{Limitations and future work.} \method works inside the bounding box of
its thresholded input plus a fixed margin, so it cannot recover anatomy missing
from that crop, and its shape representation can inherit errors of the source
models. Stronger departures from typical shape than tumor-bearing kidneys remain
untested; the change $1-\mathrm{Dice}(m,\hat m)$ between the thresholded input
mask $m$ and the corrected mask $\hat m$ can flag such cases for review without
reference annotations. Other imaging modalities, and combining the shape
representation with image features, remain to be explored.

\section{Conclusion}\label{sec:conclusion}

\method, a single text-conditioned 3D diffusion model, learns one shared shape
representation for $377$ anatomical classes from predicted segmentation
probabilities. Without receiving the image, it corrects predictions from
automatic and prompted segmentation models across $13$ public benchmarks,
transfers from CT to MRI and to four unseen lesion classes without retraining,
and produces corrected labels that train better segmentation models. Anatomical
shape can be learned once and reused across classes and imaging modalities.

\clearpage
\section*{Ethics Statement}
This work uses only publicly available, de-identified CT, CBCT and MRI datasets,
each accessed under its published license, and no new data were collected from
human subjects. \method corrects the output of segmentation models and is not
validated for clinical decision-making. Because it learns from predicted
probabilities, it can inherit errors of the source models, and its behaviour on
strongly atypical anatomy is untested (Section~\ref{sec:discussion}); the change
$1-\mathrm{Dice}(m,\hat m)$ can flag cases for human review. \dataset is derived
only from public scans, is released under the licenses of its sources, and was
audited by readers on $10{,}000$ scans (Appendix~\ref{app:downstream}).

\section*{Reproducibility Statement}
Architecture, noise schedule and training configuration are given in
Section~\ref{sec:methods} and Table~\ref{tab:config}
(Appendix~\ref{app:implementation}), and the correction procedure in
Algorithm~\ref{alg:refine} (Appendix~\ref{app:algorithm}). All training and
evaluation datasets are public and listed with their splits in
Appendices~\ref{app:data} and~\ref{app:protocol}; baseline implementations and
tuning ranges are in Appendix~\ref{app:baselines}. The code, model weights,
every anatomical description (Appendix~\ref{app:descriptions}) and the per-class
noise level $t^*$ will be released.

\section*{AI Use Statement}
We used a large language model to draft the anatomical descriptions used for
text conditioning (Section~\ref{sec:conditioning}); the authors reviewed and
edited every description against standard anatomy references. We also used a
large language model to edit the wording of this manuscript. Large language
models were not used to generate results, data or conclusions, and the authors
take full responsibility for the content.

\section*{Acknowledgments}
This work was supported by the Lustgarten Foundation for Pancreatic Cancer Research and the National Institutes of Health (NIH) under Award Number R01EB039836. We would like to thank the Johns Hopkins Research IT team in \href{https://researchit.jhu.edu/}{IT@JH} for their support and infrastructure resources where some of these analyses were conducted; especially \href{https://researchit.jhu.edu/research-hpc/}{DISCOVERY HPC}. We acknowledge CSC – IT Center for Science, Finland, for providing access to Roihu and LUMI, owned by the EuroHPC Joint Undertaking and hosted by CSC Finland in collaboration with the LUMI consortium.

\bibliography{refs,zzhou}
\bibliographystyle{iclr2027_conference}

\newpage
\appendix

\section{Training Data}\label{app:data}

\begin{figure}[htbp]
\centering
\includegraphics[width=\linewidth]{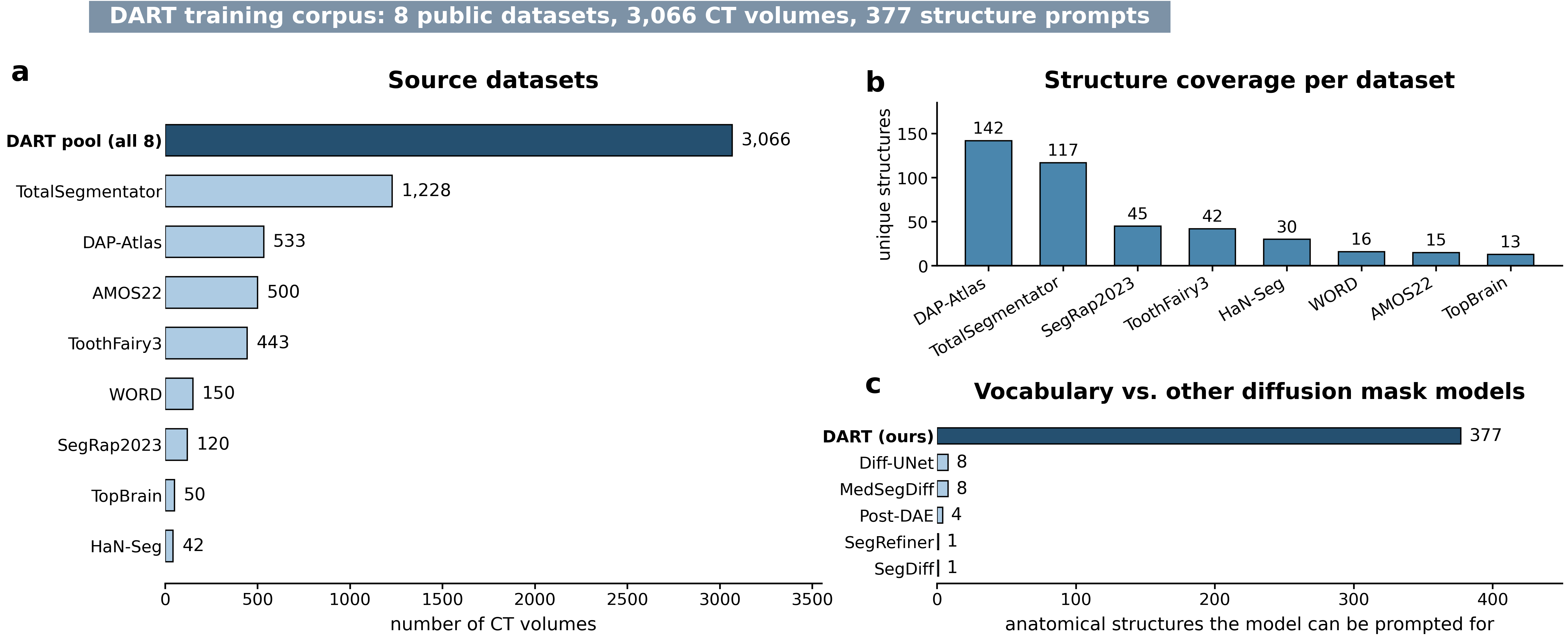}
\caption{\textbf{Composition of the training corpus.} \textbf{(a)} Volumes
contributed by each public dataset and the pooled total. \textbf{(b)} Annotated
classes per dataset. \textbf{(c)} Number of anatomical classes in the training
vocabulary, compared with the number of classes that other mask-based models
can be conditioned on; SegRefiner is
class-agnostic and Post-DAE is not a diffusion model, so this panel compares
vocabulary size only, not quality.}
\label{fig:training_data}
\end{figure}

The diffusion model is trained on segmentation probabilities predicted for
$3{,}066$ volumes from eight public datasets (Figure~\ref{fig:training_data}).
Each training example is a probability map for one anatomical class in one
volume. The source datasets' voxel-level annotations define the class inventory;
they are not the clean diffusion samples. TotalSegmentator
\citep{wasserthal2023totalsegmentator} contributes $1{,}228$ CT volumes (mean age
63.6, range 18 to 97, 41.2\% female). DAP-Atlas \citep{jaus2023dapatlas}
contributes $533$ CT volumes. AMOS22 \citep{ji2022amos} contributes its $500$ CT
volumes (mean age 53.6, range 14 to 94, 37.2\% female). ToothFairy
\citep{bolelli2024toothfairy}, in its ToothFairy3 release, contributes $443$
CBCT and CT volumes. WORD \citep{luo2021word} contributes $150$ CT volumes (mean
age 49.0, range 26 to 78, 41.3\% female). SegRap2023 \citep{luo2025segrap}
contributes $120$ CT volumes, TopBrain \citep{yang2026topbrainchallenge} $50$ CT
angiography volumes, and HaN-Seg \citep{podobnik2023hanseg} $42$ CT volumes. No
MRI or MRA volume from any dataset is used. The four lesion classes in these
datasets, both kidney cysts and the primary and nodal gross tumor volumes, are
held out. Four datasets report no demographics, and those that do skew male (55
to 63\%) and oncologic. We report demographics because a shape representation
inherits the shapes of its training cohort.

The eight datasets annotate $420$ label classes. Merging labels that refer to the
same anatomy gives a training vocabulary of $377$ anatomical classes. The number
of annotated subjects per class varies by more than two orders of magnitude, and
Appendix~\ref{app:scalingfit} shows that the model improves as a structure gains
training examples. An anatomical class is a segmentation category, so $377$
classes are not $377$ separate organs: the vocabulary includes composite tissue
and body-region classes inherited from the source datasets, such as
\texttt{muscle}, \texttt{visceral adipose tissue} and \texttt{body trunk},
alongside individual anatomical structures. The released vocabulary lists every
class with its subject count and marks the composite classes.

\section{Evaluation Protocol}\label{app:protocol}

\paragraph{Evaluation sets.} Each experiment uses a fixed evaluation set, and
sets are not pooled.
\begin{itemize}[leftmargin=1.2em,itemsep=1pt,topsep=2pt]
    \item \emph{External benchmarks} (Section~\ref{sec:beats_classical},
    Figure~\ref{fig:organ_results}, Appendix~\ref{app:external}): ten
    benchmarks from datasets that contribute nothing to training. In CT:
    FLARE22, KiTS23, MSD spleen (Task09) and MSD pancreas (Task07). In MRI: MSD
    heart (Task02), MSD hippocampus (Task04) and MSD prostate (Task05). CHAOS
    contributes three: CT, T1 MRI and T2 MRI. Inputs come from TotalSegmentator,
    AtlasNet and WholeBodySeg on CT and from TotalSegmentator MRI on MRI.
    \item \emph{Method comparison} (Section~\ref{sec:beats_classical},
    Table~\ref{tab:headline}, Appendix~\ref{app:pergroup}): $90$
    TotalSegmentator anatomical classes on which all twelve methods were run,
    all starting from MedSAM predictions. The two ASM variants produce a result for $41$ of them.
    \item \emph{Held-out WORD and AMOS22 cases} (Appendix~\ref{app:transfer}):
    TotalSegmentator and AtlasNet predictions on held-out subject-level test
    splits, $120$ cases, $14$ abdominal classes; WholeBodySeg predictions on
    held-out TotalSegmentator test cases, $104$ predicted classes.
    \item \emph{Description sensitivity} (Appendix~\ref{app:prompts}): the WORD
    and AMOS22 cases and inputs above.
    \item \emph{Dataset-level means} (Appendix~\ref{app:datasetlevel}): $77$
    anatomical classes on the ToothFairy test split and $145$ on the DAP-Atlas
    test split.
    \item \emph{Surface distance and connected components}
    (Appendix~\ref{app:datasetlevel}): held-out WORD cases, which provide dense
    abdominal annotations at native spacing, with MedSAM input.
    \item \emph{Identity controls} (Appendix~\ref{app:identity}): the ToothFairy
    test split, the held-out WORD, AMOS22 and TotalSegmentator cases above, and
    the external benchmarks.
    \item \emph{Pure-noise control} (Appendix~\ref{app:negative}): about $50$
    cases from MSD heart and FLARE22 with MedSAM input.
    \item \emph{Abnormal kidneys} (Section~\ref{sec:beats_classical},
    Appendix~\ref{app:external}): the abnormal cases of KiTS23, kidneys that
    contain tumors, with TotalSegmentator input.
    \item \emph{MRI and lesion transfer tasks} (Section~\ref{sec:scale_modality}):
    TotalSegmentator MRI predictions on MSD heart, hippocampus and prostate and on
    the PanSegNet MRI dataset \citep{zhang2025pansegnet}; MedSAM input on the
    seven transfer tasks of Figure~\ref{fig:zeroshot}a, namely the MSD prostate
    zone, hippocampus, left atrium and brain tumor, the hippocampal tumor task,
    the pancreatic cysts of Cyst-X \citep{pan2025cystx} and, in CT, the
    pancreatic lesions of PanTS \citep{li2025pants}.
    \item \emph{Downstream training} (Section~\ref{sec:downstream},
    Appendix~\ref{app:architectures}): original labels predicted by
    TotalSegmentator, AtlasNet and WholeBodySeg, corrected labels produced by
    \method from the same predictions, and held-out reference annotations for
    evaluation.
    \item \emph{Data scaling} (Appendix~\ref{app:scalingfit}): $83$ anatomical
    classes.
    \item \emph{Vocabulary size and text versus index}
    (Appendix~\ref{app:ablations}): the same $500$-case subset of
    TotalSegmentator, AMOS22 and WORD, so these two experiments are comparable
    with each other but not with the sets above.
\end{itemize}

\paragraph{Reference annotations.} Reference annotations are the released
annotations of each dataset. DAP-Atlas annotations were created by an automated
pipeline and approved through expert review \citep{jaus2023dapatlas}.

\paragraph{Bounding-box prompts for MedSAM.} MedSAM is prompted slice by slice
with 2D bounding boxes derived from the reference annotation, so the correction
methods start from predictions in the right location and the comparison measures
shape. \method never sees these bounding-box prompts. It receives MedSAM's
segmentation probabilities, thresholds them at $0.5$ to locate its crop, and
processes the original probabilities inside the crop, exactly as it does for
automatic segmentation models. Because the prompts come from the reference, the
extent of this input, and so of \method's crop, is close to the reference extent.
This favours every method that works within its input's extent, \method
included; the automatic inputs of Appendices~\ref{app:external}
and~\ref{app:transfer} carry no such information.

\paragraph{Noise level.} For each training class, $t^*$ is selected on a
validation split and then fixed. For a class outside the training vocabulary,
$t^*$ is copied from the structure in which it occurs. No test labels are
consulted, and $t^*$ is not re-selected for any transfer experiment.

\paragraph{``Up to'' values.} Unless stated otherwise, ``up to'' gives the
largest per-case gain in the stated evaluation set. Three exceptions are stated
where they occur: for the baseline comparison it gives the largest per-class
margin over a baseline (Table~\ref{tab:allstructures}), for the segmentation
models of Section~\ref{sec:downstream} the largest gain of a single
configuration, and for the lesion classes the largest gain in
mean Dice of a single task.

\paragraph{Statistics.} For each class in the $90$-class comparison, we compute
the per-case paired gain (after \method minus input) and test the null
hypothesis of zero mean gain, controlling the false discovery rate at $5\%$ with
the Benjamini--Hochberg procedure. Per-class gains are reported as mean $\pm$
standard deviation of the per-case paired gain.

\section{External Validation}\label{app:external}

\begin{figure}[htbp]
    \centering
    \includegraphics[width=0.88\linewidth]{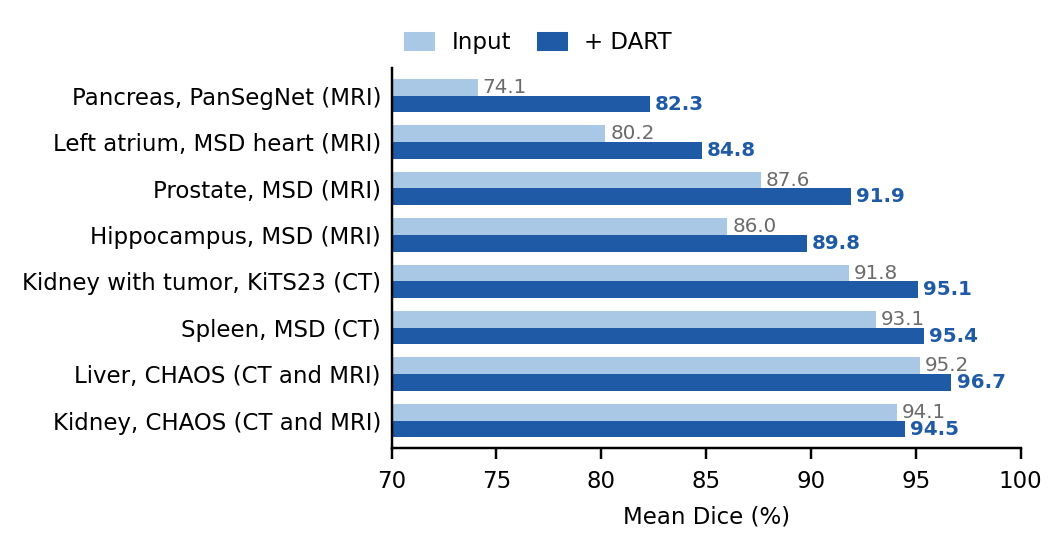}
    \caption{\textbf{\method raises mean Dice on every external benchmark.} Mean
    Dice of the input (light) and after \method (dark). The axis starts at $70$.
    CT inputs are the mean over TotalSegmentator, AtlasNet and WholeBodySeg,
    except KiTS23, which uses TotalSegmentator on its abnormal cases. MRI inputs
    come from TotalSegmentator MRI, and CHAOS averages its CT and MRI scans.
    FLARE22 and MSD pancreas appear in Figure~\ref{fig:organ_results}.}
    \label{fig:external}
\end{figure}

Figure~\ref{fig:organ_results} covers five hard structures. This section
evaluates \method more broadly, on the ten external benchmarks and on the
PanSegNet MRI dataset, none of which contributes training volumes to \method.
Every $t^*$ is fixed in advance, nothing is retuned, and every available case is
used (Figure~\ref{fig:external}).

\paragraph{Setup.} On CT we run TotalSegmentator
\citep{wasserthal2023totalsegmentator}, AtlasNet and WholeBodySeg with their
released weights and default settings, and report the mean over the three.
AtlasNet is the nnU-Net-based segmentation model released with AbdomenAtlas and
used as the segmentation stage of RadGPT \citep{bassi2025radgpt,
li2024abdomenatlas, isensee2021nnunet}. On MRI we run TotalSegmentator MRI
\citep{dantonoli2025totalsegmentatormri}. For CHAOS, Figure~\ref{fig:external}
averages its CT scans, predicted by the three CT models, with its T1 and T2 MRI
scans, predicted by TotalSegmentator MRI. Each model is evaluated on the
structures that it predicts and the benchmark annotates. The baselines of
Appendix~\ref{app:baselines} were run on the benchmarks of
Figure~\ref{fig:organ_results}; for the others we report the input, \method and,
where we ran it, the identity control. AbdomenAtlas and TotalSegmentator draw on
public CT collections, so some external images may be in an input model's
training data. \method's own training sources exclude these external cases
(Appendix~\ref{app:data}), but its inputs depend on the upstream models, so such
overlap can influence both input quality and the measured correction.

\paragraph{CT.} Mean Dice rises from $93.1$ to $95.4$ on MSD spleen, from $95.2$
to $96.7$ on the CHAOS liver, and from $94.1$ to $94.5$ on the CHAOS kidney, the
smallest gain in this section, from an input already above $94$. On the
benchmarks of Figure~\ref{fig:organ_results}, \method raises Dice by up to $31$
points on a single case, the gallbladder in FLARE22 with AtlasNet input, and
outperforms the two displayed baselines on every metric for the five structures
shown.

\paragraph{Abnormal kidneys.} In KiTS23 \citep{heller2023kits21}, the reference
kidney annotation includes the tumor. On the abnormal cases, with
TotalSegmentator input, \method raises mean Dice from $91.8$ to $95.1$ and keeps
the tumor voxels inside the kidney: the model does not erase the tumor to make
the kidney look typical. Giving the kidney the description of a different
structure lowers Dice by $4.9$ points relative to the matching description, so
the result depends on what the description names.

\paragraph{MRI.} The model is trained only on CT, and every $t^*$ is the value
selected on the CT validation split. On predictions from TotalSegmentator MRI,
mean Dice rises on all four MRI-only datasets: on the pancreas of the PanSegNet
MRI dataset \citep{zhang2025pansegnet} from $74.1$ to $82.3$, the largest gain in
this section; on the left atrium (MSD heart) from $80.2$ to $84.8$; on the
prostate from $87.6$ to $91.9$; and on the hippocampus, averaged over left and
right, from $86.0$ to $89.8$. On single prostate cases the gain reaches $10$ Dice
points and $28$ NSD points at $2$\,mm. The MSD heart and hippocampus also appear
in Figure~\ref{fig:zeroshot}a with MedSAM input, so the numbers there differ.

\paragraph{Identity control.} Running \method's crop, resize and threshold steps
without the diffusion model adds $0.02$ Dice points on the spleen, $0.09$ on the
CHAOS liver, $0.21$ on the left atrium and $0.24$ on the hippocampus, between
$0.9\%$ and $6.2\%$ of \method's gain on the same benchmark.

\paragraph{Summary.} Mean Dice rises on every benchmark in this section, and the
gain is broadly largest where the input is weakest: $8.2$ points on an input of
$74.1$, against $0.4$ points on an input above $94$. Held-out WORD and AMOS22
cases show the same pattern (Appendix~\ref{app:transfer}).

\section{Automatic Segmentation Models on Held-out WORD and AMOS22 Cases}\label{app:transfer}

Here we evaluate automatic segmentation models on held-out cases from two
datasets that also contribute training volumes, which lets us compare two models
on the same cases.

\paragraph{Setup.} We run TotalSegmentator and AtlasNet on held-out
subject-level test splits of WORD \citep{luo2021word} and AMOS22
\citep{ji2022amos}, $120$ cases in total, none of which contributed probability
maps to \method's training. We report the $14$ abdominal classes the two models
share with those reference annotations. On TotalSegmentator inputs we exclude the
rectum, which that model does not segment as its own class (input Dice $11.6$);
it is kept for AtlasNet.

\paragraph{Two segmentation models, two starting points.} AbdomenAtlas includes
WORD and AMOS22 among its sources, so AtlasNet starts from stronger inputs than
TotalSegmentator on most structures, and \method gains correspondingly less. The
pair therefore shows how the gain depends on input quality.

\begin{table}[htbp]
\centering\small
\setlength{\tabcolsep}{5pt}
\caption{\textbf{\method on TotalSegmentator predictions.} Dice before and after
\method on held-out WORD and AMOS22 cases, sorted by gain. $\pm$ is the standard
deviation of the per-case paired gain. Mean Dice improves for every evaluated
class.}
\label{tab:transfer_ts}
\begin{tabular}{@{}lccc@{}}
\toprule
Anatomical class & TotalSegmentator & + \method & Gain \\
\midrule
Adrenal glands & 64.70 & \best{70.46} & $+5.76\pm3.14$ \\
Duodenum       & 60.44 & \best{64.50} & $+4.07\pm2.40$ \\
Esophagus      & 76.51 & \best{80.18} & $+3.67\pm2.51$ \\
Gallbladder    & 66.85 & \best{70.20} & $+3.35\pm3.92$ \\
Colon          & 76.43 & \best{79.54} & $+3.11\pm1.72$ \\
Pancreas       & 79.67 & \best{82.74} & $+3.08\pm2.28$ \\
Intestine      & 80.92 & \best{83.85} & $+2.93\pm3.51$ \\
Kidney (left)  & 93.44 & \best{95.08} & $+1.65\pm1.19$ \\
Kidney (right) & 93.53 & \best{95.11} & $+1.58\pm1.37$ \\
Spleen         & 94.42 & \best{95.63} & $+1.21\pm0.80$ \\
Stomach        & 92.22 & \best{93.06} & $+0.83\pm0.51$ \\
Bladder        & 92.09 & \best{92.62} & $+0.53\pm0.50$ \\
Liver          & 95.74 & \best{96.01} & $+0.28\pm0.47$ \\
\midrule
\textbf{Mean} & 82.07 & \textbf{84.54} & $\mathbf{+2.47}$ \\
\bottomrule
\end{tabular}
\end{table}

\begin{table}[htbp]
\centering\small
\setlength{\tabcolsep}{5pt}
\caption{\textbf{\method on AtlasNet predictions.} Same protocol as
Table~\ref{tab:transfer_ts}. Mean Dice improves for thirteen of fourteen classes;
the intestine loses $0.16$ points, which is within the per-case spread.}
\label{tab:transfer_aa}
\begin{tabular}{@{}lccc@{}}
\toprule
Anatomical class & AtlasNet & + \method & Gain \\
\midrule
Rectum         & 72.13 & \best{75.64} & $+3.51\pm4.56$ \\
Adrenal glands & 69.43 & \best{72.88} & $+3.45\pm2.84$ \\
Esophagus      & 76.89 & \best{78.41} & $+1.52\pm1.97$ \\
Gallbladder    & 79.14 & \best{80.59} & $+1.44\pm1.66$ \\
Duodenum       & 72.75 & \best{73.70} & $+0.95\pm1.06$ \\
Bladder        & 85.60 & \best{86.35} & $+0.75\pm0.95$ \\
Pancreas       & 82.82 & \best{83.55} & $+0.73\pm1.28$ \\
Liver          & 95.33 & \best{96.06} & $+0.73\pm0.69$ \\
Spleen         & 95.15 & \best{95.62} & $+0.47\pm0.66$ \\
Kidney (right) & 93.31 & \best{93.75} & $+0.44\pm0.65$ \\
Stomach        & 92.24 & \best{92.53} & $+0.29\pm0.52$ \\
Kidney (left)  & 93.96 & \best{94.25} & $+0.28\pm0.28$ \\
Colon          & 82.72 & \best{82.87} & $+0.14\pm1.39$ \\
Intestine      & \best{87.03} & 86.87 & $-0.16\pm0.78$ \\
\midrule
\textbf{Mean} & 84.18 & \textbf{85.22} & $\mathbf{+1.04}$ \\
\bottomrule
\end{tabular}
\end{table}

\begin{figure}[htbp]
\centering
\includegraphics[width=\linewidth]{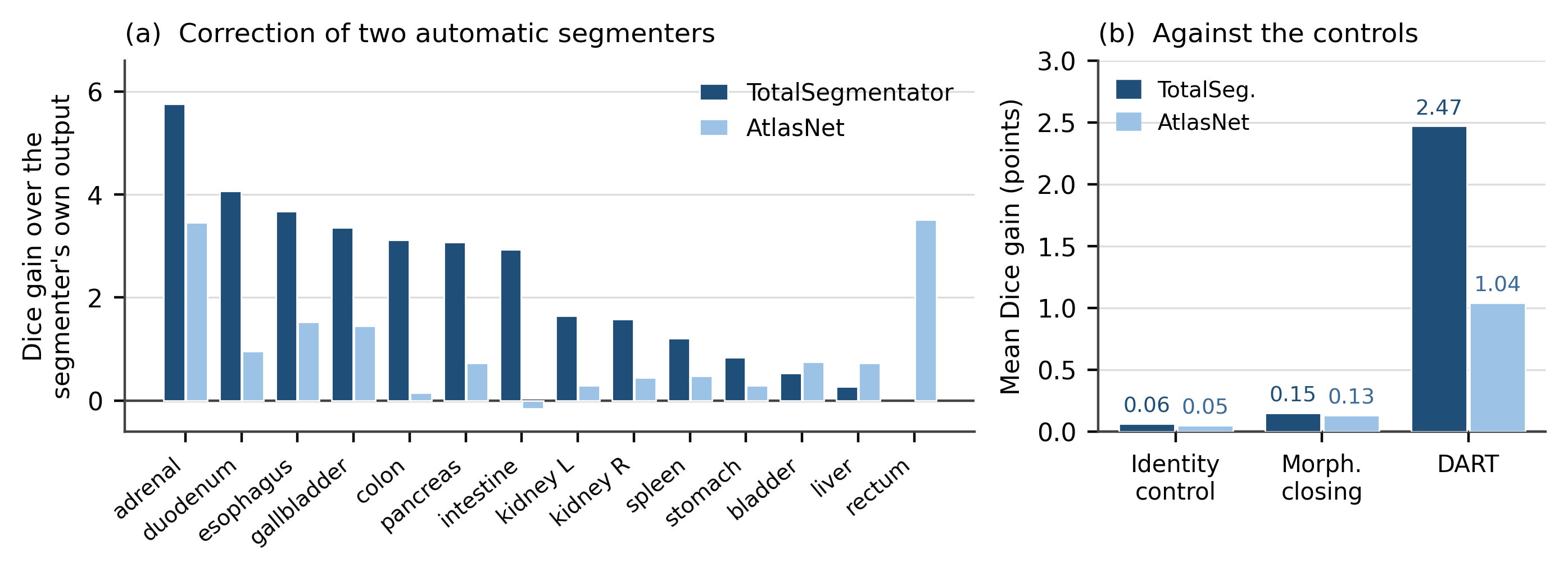}
\caption{\textbf{\method improves the predictions of two automatic segmentation
models.} \textbf{(a)} Dice gain per class over each model's own prediction, on
held-out WORD and AMOS22 cases with the per-class $t^*$ unchanged.
\textbf{(b)} Mean gain against the two controls that use the same inputs: the
identity path with the diffusion model removed, and morphological closing with
its structuring element swept per class on the validation split.}
\label{fig:transfer}
\end{figure}

\paragraph{Results.} \method improves $26$ of the $27$ class--model pairs
(Tables~\ref{tab:transfer_ts} and~\ref{tab:transfer_aa},
Figure~\ref{fig:transfer}). Mean Dice rises by $2.47$ points on TotalSegmentator
predictions and by $1.04$ on AtlasNet predictions. On the same inputs, the
identity control of Appendix~\ref{app:identity} adds $0.06$ and $0.05$ points,
and morphological closing, with its structuring element swept per class on the
validation split, adds $0.15$ and $0.13$.

\paragraph{The gain follows the headroom.} Reading the two tables together
separates the effect of the structure from the effect of the input. The same
structure gains less when the input is better: the duodenum gains $4.07$ from a
$60.4$ input and $0.95$ from a $72.8$ input; the colon gains $3.11$ from $76.4$
and $0.14$ from $82.7$; the intestine gains $2.93$ from $80.9$ and loses $0.16$
from $87.0$. Within each model, the largest gains fall on the adrenal glands,
duodenum, esophagus and gallbladder, where both models are weakest, and the
smallest on the liver, spleen, bladder and stomach, which are close to solved.
Across structures, the gain falls as the input improves, with a Spearman
correlation of $\rho=-0.89$ for TotalSegmentator ($p<0.001$) and $\rho=-0.68$
for AtlasNet ($p=0.008$). This is the behaviour a shape representation should
show.

\paragraph{A third segmentation model on a different dataset.} We also apply
\method to the predictions of WholeBodySeg, the MONAI whole-body CT
segmentation model \citep{cardoso2022monai}: a SegResNet that predicts $104$
anatomical classes, trained on a different TotalSegmentator release from the one
we test on. We run it on TotalSegmentator test cases held out from \method's
training and correct every class it predicts, with the same frozen model and
per-class $t^*$. Mean Dice rises by $3.77$ points. The identity control, run on
the same inputs, adds $0.08$ points, $2.1\%$ of the gain, so nearly all of the
gain comes from the diffusion model. This extends the results for automatic
segmentation models to a third model, a third dataset and structures across the
whole body.

\section{Anatomical Description Sensitivity: Paraphrase and Mismatch}\label{app:prompts}

\paragraph{Setup.} We condition the frozen model on three description variants
on the WORD and AMOS22 inputs of Appendix~\ref{app:transfer}, holding the noise
level $t^*$, the random seed and everything else fixed, and report the change
in Dice relative to the matching description. \emph{Paraphrase} rewrites the
description in different words while keeping the name, shape cue and location
cue, for example replacing \textit{``pancreas, elongated lobulated gland,
retroperitoneum behind the stomach from duodenal curve to splenic hilum''} with
a rewording that names the same organ, the same shape and the same neighbours.
\emph{Mismatch} substitutes the unmodified description of a different class in
the vocabulary, chosen by hand for each class. The paraphrase run also includes
the left and right femur. Values are averaged over the segmentation models that
predict each class.

\begin{figure}[htbp]
\centering
\includegraphics[width=\linewidth]{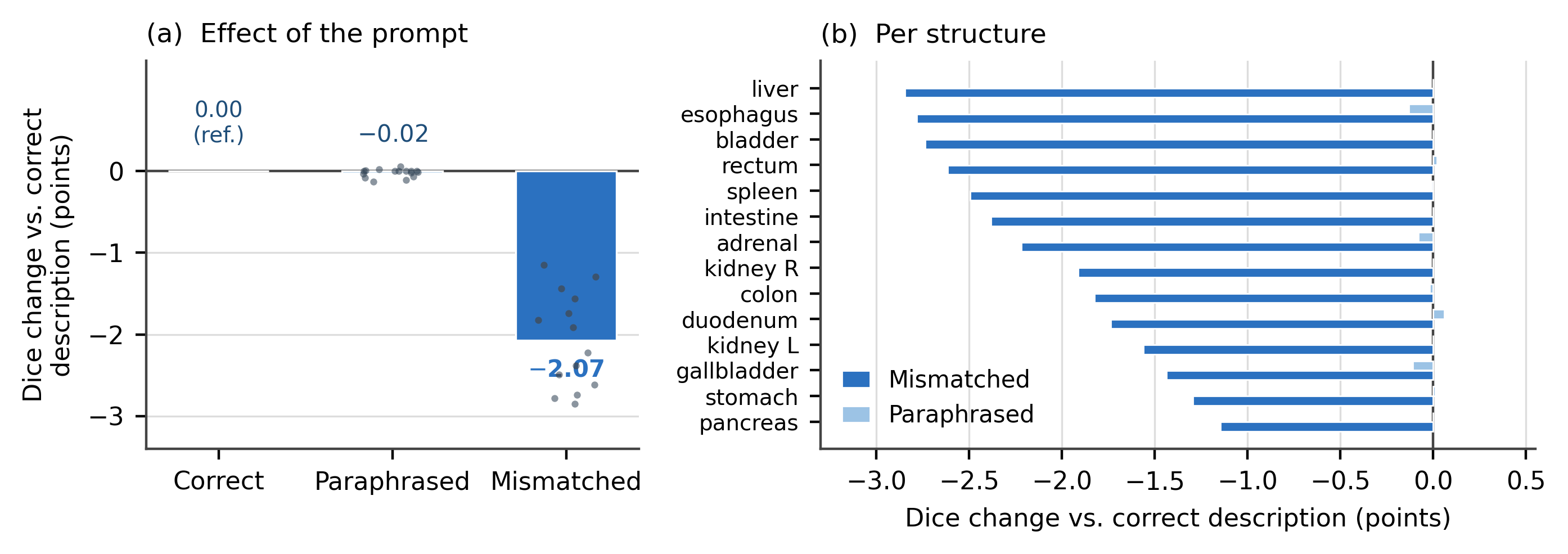}
\caption{\textbf{\method follows what the description says, not its wording.}
Change in Dice relative to the matching description, on the predictions of the
automatic segmentation models of Appendix~\ref{app:transfer}, with the noise
level and seed held fixed. \textbf{(a)} A paraphrase that keeps the anatomy but
changes the words barely changes Dice ($-0.02$ points); substituting another
structure's description costs $2.07$ points. Dots are individual anatomical
classes. \textbf{(b)} The same, per class. Every class loses Dice under a
mismatched description; none moves by more than $0.13$ points under a
paraphrase.}
\label{fig:conditioning}
\end{figure}

\paragraph{Results.} A paraphrase changes Dice by $-0.02$ points on average over
$16$ anatomical classes and by at most $0.13$ points on any one
(Figure~\ref{fig:conditioning}, Table~\ref{tab:prompts}). A mismatched
description costs $2.07$ points on average over $14$ anatomical classes and
lowers Dice on every one. The two effects differ by two orders of magnitude. A
model that treated the description as an arbitrary code would be thrown off by a
paraphrase, because a paraphrase is a different string. A model that ignored the
description would not notice a mismatch. \method is insensitive to wording and
sensitive to what the description names.

\paragraph{What the mismatch test measures.} Both variants run at the noise level
$t^*$ of the matching class, which keeps the output close to its input. The
mismatch cost measures sensitivity to the anatomical description at this noise
level; it does not isolate a distinct shape representation for each class.
Across the $14$ classes, the mismatch cost is unrelated to the gain from \method
(Spearman $\rho=-0.07$, $p=0.82$).

\begin{table}[htbp]
\centering\small
\setlength{\tabcolsep}{6pt}
\caption{\textbf{Description sensitivity.} Change in Dice (points) relative to
the matching description, averaged over the two segmentation models of
Appendix~\ref{app:transfer}. `---' means the class was not included in that
run.}
\label{tab:prompts}
\begin{tabular}{@{}lcc@{}}
\toprule
Anatomical class & Paraphrased & Mismatched \\
\midrule
Adrenal glands & $-0.08$ & $-2.22$ \\
Bladder        & $0.00$ & $-2.74$ \\
Colon          & $-0.02$ & $-1.83$ \\
Duodenum       & $+0.06$ & $-1.74$ \\
Esophagus      & $-0.13$ & $-2.78$ \\
Femur (left)   & $-0.03$ & --- \\
Femur (right)  & $-0.07$ & --- \\
Gallbladder    & $-0.11$ & $-1.44$ \\
Intestine      & $0.00$ & $-2.38$ \\
Kidney (left)  & $0.00$ & $-1.56$ \\
Kidney (right) & $0.00$ & $-1.91$ \\
Liver          & $0.00$ & $-2.85$ \\
Pancreas       & $0.00$ & $-1.15$ \\
Rectum         & $+0.02$ & $-2.61$ \\
Spleen         & $-0.01$ & $-2.49$ \\
Stomach        & $+0.01$ & $-1.29$ \\
\midrule
\textbf{Mean}  & $\mathbf{-0.02}$ & $\mathbf{-2.07}$ \\
\bottomrule
\end{tabular}
\end{table}

\section{Dataset-Level Results and Segmentation Quality}\label{app:datasetlevel}

\begin{figure}[htbp]
\centering
\includegraphics[width=\linewidth]{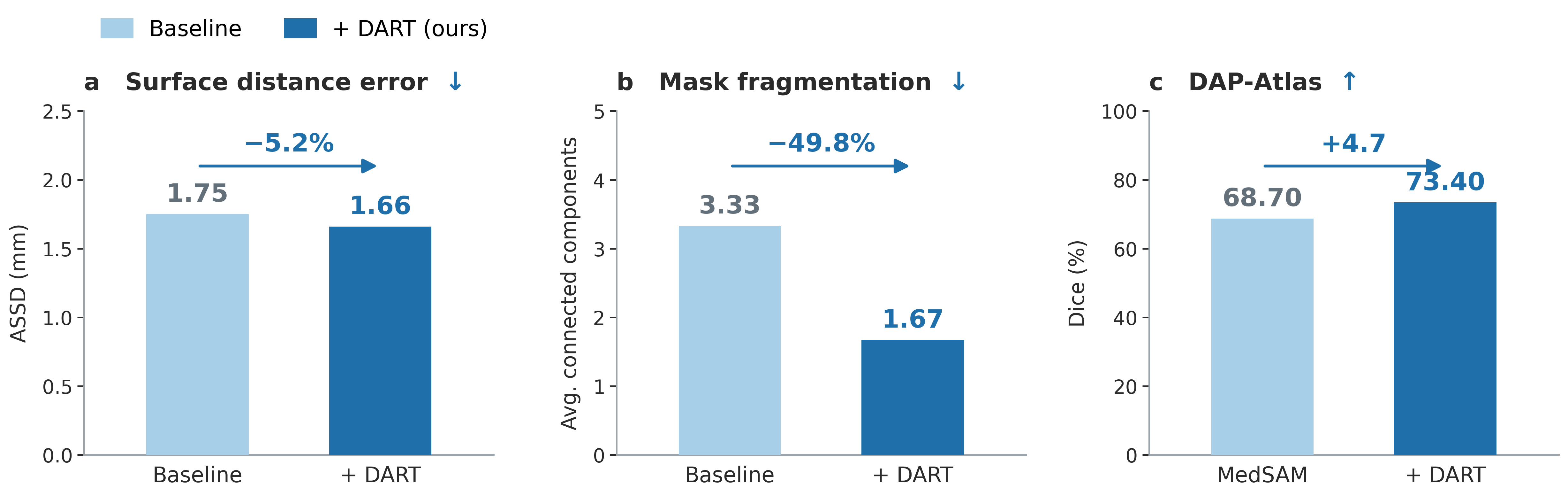}
\caption{\textbf{Boundary and fragmentation metrics.} \textbf{(a)} Average
symmetric surface distance before and after \method on held-out WORD cases.
\textbf{(b)} Connected components per mask on the same cases. \textbf{(c)} Mean
Dice on DAP-Atlas for the MedSAM input and after \method.}
\label{fig:maskquality}
\end{figure}

Over the $77$ ToothFairy anatomical classes, \method raises mean Dice from
$53.83$ to $57.67$ ($+3.8$ points). Over the $145$ DAP-Atlas anatomical classes,
it raises mean Dice from $68.70$ to $73.40$ ($+4.7$ points). Both gains are
larger than on the $90$ TotalSegmentator anatomical classes, which include organs
whose inputs are already accurate. On held-out WORD cases with MedSAM input,
\method reduces the average symmetric surface distance from $1.75$\,mm to
$1.66$\,mm ($-5.2\%$) and the number of connected components per mask from
$3.33$ to $1.67$ ($-49.8\%$) (Figure~\ref{fig:maskquality}).

\section{Identity Controls}\label{app:identity}

\method locates its crop by thresholding the input probabilities at $0.5$,
crops and resizes the original probabilities to \cropvol, runs the reverse
process, resamples back to the native crop by trilinear interpolation, and
thresholds the result. Resampling and thresholding can change Dice on their own,
so two controls run this path without the diffusion model. \emph{Identity}
crops, resizes, resamples and thresholds at $0.5$, with no noising or denoising.
\emph{Identity (volume-matched)} does the same but chooses the threshold that
reproduces the foreground voxel count of the input probabilities thresholded at
$0.5$, which removes any volume bias of a fixed threshold.

On ToothFairy, against a MedSAM input of $53.83$ Dice, the identity control
reaches $53.96$ and the volume-matched identity control reaches $54.00$, while
\method reaches $57.67$. The resampling path therefore accounts for $3.4\%$ of
the gain, and the volume bias for $4.4\%$. On the same split, Gaussian smoothing
reaches $54.12$, a median filter $54.23$, the convex hull $54.69$, morphological
closing $55.72$, and convex hull followed by closing $56.18$. \method thus
exceeds the strongest traditional method by $1.5$ points and the identity
control by $3.7$ points. Across this evaluation set, the gain is positive on
$652$ of $658$ class--case pairs.

On the held-out WORD and AMOS22 inputs of Appendix~\ref{app:transfer}, the same
identity control adds $0.06$ Dice points on TotalSegmentator predictions and
$0.05$ on AtlasNet predictions, and on held-out TotalSegmentator cases it adds
$0.08$ on WholeBodySeg predictions, against $2.47$, $1.04$ and $3.77$ for
\method. On the external benchmarks of Appendix~\ref{app:external}, it adds
between $0.02$ and $0.24$ points, $0.9\%$ to $6.2\%$ of the gain. On the
benchmarks of Figure~\ref{fig:organ_results}, it is included as a baseline; on
the left adrenal gland it is among the two strongest, because most other
baselines lower Dice there.

\section{Additional Observations}\label{app:negative}

\paragraph{The opposite control: sampling from pure noise.} The identity
control removes the diffusion model; this control removes the input. The reverse
process starts from pure noise at $t=T$, in the same crop and under the same
description, so the output depends on the description and the crop but not on
the probability values within that crop. On about $50$ cases from MSD heart
(MRI) and FLARE22 (CT) with MedSAM input, the input scores $77.37$ Dice, \method
$78.91$, and sampling from pure noise $34.08$, or $34.31$ with the threshold
chosen to maximise Dice. Because MedSAM is prompted with bounding boxes from the
reference annotation, the crop nearly matches the reference extent; even so, the
description alone recovers less than half of the input's Dice. \method therefore
edits its input rather than replacing it with a shape generated for the class.

\paragraph{The baselines are weak on the TotalSegmentator benchmark.} Nine of the
eleven baselines lower Dice on more than half of the classes they run on
(Table~\ref{tab:allstructures}): the convex hull on $65$ of $90$, SegRefiner on
$87$ and Post-DAE on all $90$, while the two classical methods that read the
image lose $21.6$ and $28.5$ points on average and the two ASM variants about
$59$. We re-checked each implementation after observing this. The losses follow
from what each method assumes, as the paragraphs below explain, not from
implementation errors.

\paragraph{Fixed rules help some structures and hurt others.} The convex hull
improves the left autochthon by $2.8$ Dice points but lowers the pancreas by
$10.8$ and the colon by $7.1$: a convex shape suits compact structures and not
lobulated or tubular ones. Hole filling barely changes Dice, so the input errors
lie at boundaries and in connectivity rather than in interior holes.

\paragraph{Methods that read the image are unsafe to apply blindly.} GMM
graph-cut and geodesic active contours work where a strong intensity edge
coincides with the anatomical boundary and fail elsewhere. Over the $90$
anatomical classes, their mean gains are $-21.6$ and $-28.5$ Dice points. Neither
can be applied to an unknown structure without first deciding whether it will
help.

\paragraph{Classes without a significant gain.} At a $5\%$ false discovery rate,
the gain is not significant on six of the $90$ anatomical classes: liver
($+0.01$ points), spleen ($+0.09$), left scapula ($+0.15$), right hip ($+0.18$),
right kidney ($+0.18$) and left kidney ($+0.30$). The four organs have inputs
above $94$ Dice and leave little to improve. The two bones have low inputs
($51.0$ and $59.9$ Dice), yet the model changes them little.

\paragraph{How the gain relates to input quality depends on the input.} On the
fixed MedSAM input, regressing the per-class gain on input Dice over the $90$
anatomical classes gives a slope of $-0.003$ ($r=-0.03$, $p=0.78$), and grouping
is more informative than the regression: $+2.5$ points on the $59$ anatomical
classes below $80$ Dice and $+0.8$ on the nine above $90$. We attribute the weak
relationship to the input itself: bounding-box prompts from the reference
annotation raise input Dice unevenly across structures and compress the range,
so input quality is a poor proxy for how much shape error remains. On
predictions from automatic segmentation models in Appendix~\ref{app:transfer},
which receive no such prompts, the relationship is clear: the gain falls as the
input improves ($\rho=-0.89$ and $-0.68$), both across structures and for the
same structure across the two models. The second setting is the one that matches
deployment.

\paragraph{ASM depends strongly on anatomy.} On the $41$ TotalSegmentator
anatomical classes where they produce a result, both ASM variants lower mean Dice
by about $59$ points. A linear shape space fitted to a few dozen aligned examples
cannot represent the variability of hollow and branching structures and pulls the
mask toward the mean shape. For thin structures such as the iliac arteries and
the subclavian vessels, a small misalignment between that mean shape and the true
structure leaves almost no overlap. The two variants differ on only $3$ of these
$41$ anatomical classes. For example, on the right iliac vein, \method reaches
$87.0$ Dice and ASM $6.1$, a difference of $80.9$ points. The input is already
$84.7$, and morphological closing, the strongest baseline for this class,
reaches $86.1$. Thus, the $80.9$-point difference reflects an ASM failure;
\method improves the input by $2.3$ points and exceeds the strongest baseline by
$0.9$.

\paragraph{Why the learned baselines fall short.} All three are trained on
segmentation predictions paired with their reference annotations, so they learn
from the kind of errors they are tested on. Post-DAE must reconstruct the whole
mask through a compact code, which likely costs detail even where the input is
already right; it lowers Dice on every class. SegRefiner reads the image, but it
was designed for 2D natural images; applied slice by slice, it lacks 3D context
and lowers Dice on $87$ of the $90$ classes. The single-pass U-Net shares
\method's backbone and text conditioning but maps its input to the reference
annotation in one deterministic forward pass, without diffusion or partial
noising. Trained to map a mask that is already close to its reference onto that
reference in one step, it learns little more than the identity: it stays within
$1.6$ Dice points of its input on every class, and on the right adrenal gland it
returns its input unchanged on every case (per-case gain $0.0\pm0.0$ in
Table~\ref{tab:headline}). \method instead adds noise first, which turns the input
into a noisy sample of the form seen in training and lets the model make larger
changes of shape.

\section{Scaling with Training Examples}\label{app:scalingfit}

Figure~\ref{fig:scaling} (Section~\ref{sec:scale_class}) retrains the model on nested subsets with $10\%$ to
$100\%$ of each structure's training probability maps, keeping everything else
fixed, and evaluates $83$ anatomical classes. This experiment uses its own set of
classes and its own starting point, separate from Table~\ref{tab:allstructures}.

Grouped by anatomy (Figure~\ref{fig:scaling}b), skeletal structures saturate
earliest, viscera reach the highest ceiling and vessels the lowest. We see two
reasons for the low-data points, where the output falls below the
input. With few training examples, the learned distribution is a coarse model of
the structure. And $t^*$ is held at the values chosen for the full model, so a
weak model may change a prediction as much as a strong one. Together they can
pull a mask toward an inaccurate shape. With more training examples, the learned
shape representation improves, and correction at the same noise level becomes
useful.

The mean curve is well described by a saturating power law in the percentage
$f$ of training examples, with Dice on a $0$--$1$ scale,
$D(f) = D_\infty - a f^{-\alpha}$, with $D_\infty = 0.65$, $a = 2.63$ and
$\alpha = 0.83$. Setting $D(f)$ to the mean input Dice of $0.53$ gives
$f \approx 41\%$. We use the fit as a summary of the measured curve. The
conclusions rest on the measured points themselves: mean Dice rises
monotonically, is still rising between $75\%$ and $100\%$ of the training
examples, and exceeds the input from about $40\%$ onward.

\section{Vocabulary and Conditioning Ablations}\label{app:ablations}

Both experiments use the same $500$-case subset of TotalSegmentator, AMOS22 and
WORD, with subject-level disjoint splits. Absolute Dice values are therefore not
comparable to the $90$-class tables or to Appendix~\ref{app:transfer}.

\begin{table}[htbp]
\centering\small
\caption{\textbf{Expanding the vocabulary preserves mean Dice on existing
classes.} Mean Dice on the same $50$ anatomical classes, on TotalSegmentator,
AMOS22 and WORD test cases, for models trained on $50$ anatomical classes and on
the full vocabulary.}
\label{tab:structure_scaling}
\begin{tabular}{@{}lcc@{}}
\toprule
Training vocabulary & Test classes & Dice \\
\midrule
50 & 50 & 73.85 \\
377 & 50 & \textbf{74.20} \\
\bottomrule
\end{tabular}
\end{table}

\begin{table}[htbp]
\centering\small
\caption{\textbf{Anatomical descriptions outperform class indices.} Replacing the
text embedding with a learned class embedding indexed by an integer class
identifier, with everything else fixed, reduces the gain from $+5.49$ to $+0.39$
Dice points on TotalSegmentator, AMOS22 and WORD test cases. The evaluation set
includes classes outside the index-conditioned model's vocabulary; see the text
below.}
\label{tab:indexing_comparison}
\begin{tabular}{@{}lcc@{}}
\toprule
Model & Dice & Gain \\
\midrule
Input (before \method) & 68.71 & --- \\
Index-conditioned & 69.10 & $+0.39$ \\
Text-conditioned (\method) & \textbf{74.20} & $\mathbf{+5.49}$ \\
\bottomrule
\end{tabular}
\end{table}

\paragraph{What the vocabulary comparison measures.} Table~\ref{tab:structure_scaling}
compares two separately trained models with the same architecture on the same
$50$ classes. The $377$-class model also covers $327$ additional classes, so the
comparison tests whether sharing one shape representation across many more
classes lowers mean Dice on the original ones; it does not.

\paragraph{What the index comparison measures.} An index-conditioned model
learns a separate class embedding for each class from its training examples. The
source datasets' annotation counts vary by more than two orders of magnitude
across classes (Appendix~\ref{app:data}), so class coverage is uneven, which may
limit embeddings for classes with few training examples. Text embeddings can
instead encode similarities between anatomical descriptions. However, the
evaluation also includes classes outside the index-conditioned model's
vocabulary, which an index cannot represent. Table~\ref{tab:indexing_comparison}
therefore combines differences in conditioning with differences in vocabulary
coverage. It does not establish how much of the gap is due to each factor.

The index comparison involves two separately trained models, while the mismatch
test of Appendix~\ref{app:prompts} keeps one model and changes only the
anatomical description. This is why the text--index gap ($5.10$ points) and the
mismatch cost ($2.07$ points) measure different things: the first compares two
models, the second changes the anatomical description supplied to the same
model.

\section{\method Algorithm}\label{app:algorithm}

Algorithm~\ref{alg:refine} gives the full procedure of
Section~\ref{sec:refinement}. It receives a segmentation probability map, an
anatomical description and the per-class noise level. The binary mask $m$
locates the crop; the diffusion input retains the original probabilities.

\begin{algorithm}[htbp]
\caption{\method}\label{alg:refine}
\begin{algorithmic}[1]
\Require segmentation probability map $p$; anatomical description $s$; noise level $t^*$
\Ensure corrected segmentation mask in the native geometry of $p$
\State $m \gets \mathbf{1}[p>0.5]$ \Comment{Localization only}
\If{$m$ is empty} \State \Return $m$ \EndIf
\State $B \gets \mathrm{BoundingBox}(m)$ with the fixed margin
\State $c \gets g(E(s))$ \Comment{Eq.~\ref{eq:conditioning}}
\State $x_0^{\mathrm{in}} \gets \mathrm{Resize}(\mathrm{Crop}(p,B), \cropsize^3)$
\State $\epsilon \sim \mathcal{N}(0,\mathbf{I})$
\State $x_{t^*} \gets \sqrt{\bar\alpha_{t^*}}\,x_0^{\mathrm{in}} + \sqrt{1-\bar\alpha_{t^*}}\,\epsilon$ \Comment{Eq.~\ref{eq:forward}}
\For{$t = t^*, t^*-1, \ldots, 1$}
  \State $x_{t-1} \gets \mathrm{DDPMStep}(x_t, \epsilon_\theta(x_t,t,c))$
\EndFor
\State \Return $\mathrm{Uncrop}(\mathrm{Resample}(x_0) > 0.5,B)$
\end{algorithmic}
\end{algorithm}

\section{Implementation Details}\label{app:implementation}

\begin{figure}[htbp]
    \centering
    \includegraphics[width=\linewidth]{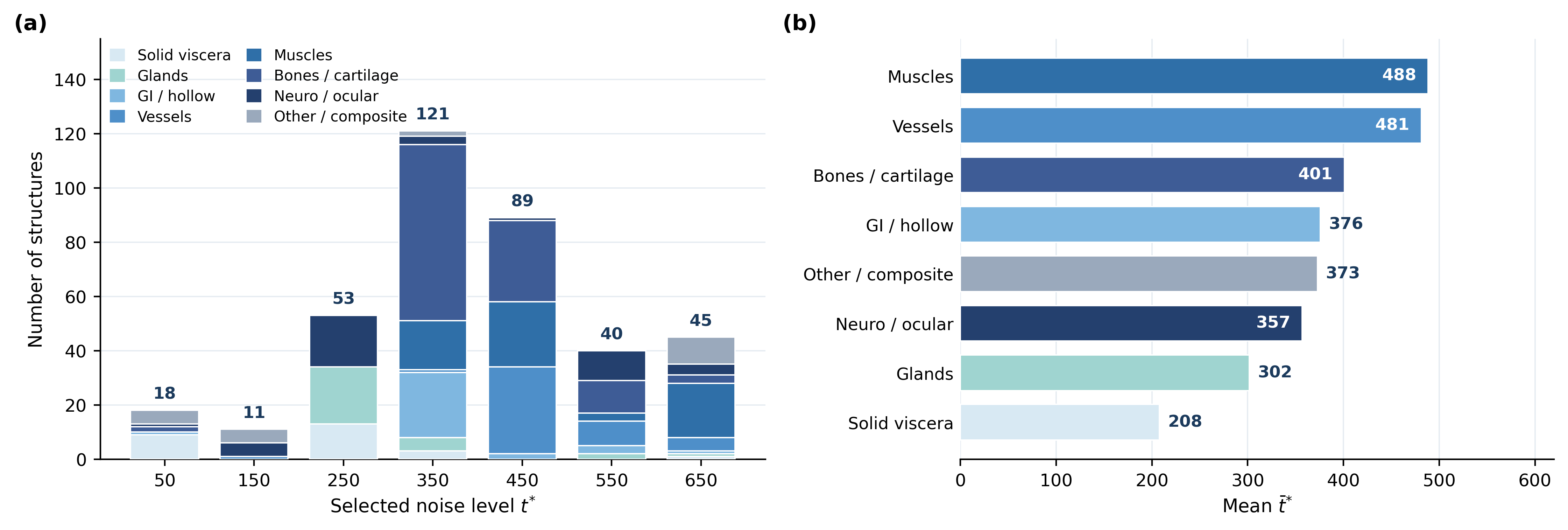}
    \caption{\textbf{Selected noise levels follow anatomy.} \textbf{(a)}
    Validation-selected $t^*$ for the training vocabulary, stacked by
    anatomical group. \textbf{(b)} Mean $t^*$ per group. Compact solid organs
    need the least noise; muscles and vessels need the most. The per-class
    values are released with the code.}
    \label{fig:noise}
\end{figure}

Table~\ref{tab:config} lists the training and inference configuration, and
Figure~\ref{fig:noise} shows the validation-selected noise levels.

\begin{table}[htbp]
\centering\small
\caption{\textbf{Training and inference configuration.}}
\label{tab:config}
\begin{tabular}{@{}ll@{}}
\toprule
\multicolumn{2}{@{}l}{\textit{Architecture}} \\
Backbone & 3D MONAI \texttt{DiffusionModelUNet} \\
Resolution levels & 4, with channels $(16,32,64,128)$ \\
Input & 1-channel probability map, cropped and resized to \cropvol \\
\midrule
\multicolumn{2}{@{}l}{\textit{Text conditioning}} \\
Text encoder & \texttt{all-MiniLM-L6-v2}, frozen \\
Text embedding & 384-dim, computed once and stored \\
Projection & Linear(384,128), SiLU, Linear(128,$d$), trainable \\
Injection & added to the timestep embedding in every residual block \\
\midrule
\multicolumn{2}{@{}l}{\textit{Diffusion}} \\
Total steps $T$ & 1000 \\
Noise schedule & linear, $\beta_1 = 1{\times}10^{-4}$, $\beta_T = 0.02$ \\
Objective & noise prediction with mean squared error \\
Sampler & DDPM ancestral sampling from $t^*$ to 0 \\
Noise level $t^*$ & per class, from $\{50,150,\dots,650\}$ \\
\midrule
\multicolumn{2}{@{}l}{\textit{Optimization}} \\
Optimizer & AdamW, learning rate $2{\times}10^{-4}$, weight decay $1{\times}10^{-5}$ \\
Schedule & cosine annealing \\
Batch size / epochs & 24 / 113 \\
Gradient clipping & global norm 1.0 \\
Precision & automatic mixed precision \\
Augmentation & random flips along all axes, random $90^\circ$ rotations \\
\bottomrule
\end{tabular}
\end{table}

\paragraph{Anatomical descriptions.} Each description should explain the target
anatomy's shape, location and distinguishing characteristics; its level of detail
depends on the anatomy. The implementation processes descriptions within the
input limit of the chosen text encoder.

\paragraph{Inference cost.} \method takes $0.59$\,s per probability map on one
AMD MI250X GPU at \cropvol. A scan containing $50$ anatomical classes therefore
costs about half a minute of GPU time on top of segmentation. \method runs once
over each predicted probability map.

\subsection{Segmentation models for the downstream experiment}\label{app:architectures}

Table~\ref{tab:longer} lists the fifteen segmentation model configurations. They
cover convolutional (U-Net \citep{ronneberger2015unet}, Basic U-Net
\citep{falk2019u} and V-Net), residual (Residual U-Net and SegResNet
\citep{myronenko2018segresnet}), attention-gated (Attention U-Net
\citep{oktay2018attention}), self-configuring (nnU-Net \citep{isensee2021nnunet}
in its DynUNet implementation) and transformer (UNETR) architectures, in several
widths and depths. The original-label arm trains on masks predicted by a combination of
three automatic segmentation models, TotalSegmentator
\citep{wasserthal2023totalsegmentator}, AtlasNet \citep{bassi2025radgpt,
li2024abdomenatlas} and WholeBodySeg \citep{cardoso2022monai}, and the
corrected-label arm on the masks that \method produces from the same probability
maps. Both arms of each configuration share one training
schedule and are evaluated against the same held-out reference annotations, so
they differ only in their training labels.

Corrected labels improve every configuration. We summarize the eleven whose
original-label arm reaches $45$ Dice, which train well under this schedule:
gains range from $0.8$ to $10.4$ points, with a median of $3.9$ and a mean of
$4.9$. Figure~\ref{fig:zeroshot}b shows seven of these configurations; its
nnU-Net arm is the five-level DynUNet.

\begin{table}[htbp]
\centering
\small
\caption{\textbf{Corrected labels improve all fifteen segmentation model
configurations.} Dice of each configuration trained on the original labels and on
the same labels after \method, with one shared training schedule and every other
factor fixed.
Configurations are grouped by whether the original-label arm reaches $45$
Dice.}
\label{tab:longer}
\begin{tabular}{@{}lccc@{}}
\toprule
Configuration & Original labels & Corrected labels & Gain \\
\midrule
\multicolumn{4}{@{}l}{\textit{Original-label arm at least 45 Dice}} \\
Attention U-Net (wide) & 62.3 & \textbf{72.7} & $+10.4$ \\
Attention U-Net & 57.1 & \textbf{67.5} & $+10.4$ \\
Residual U-Net & 57.2 & \textbf{66.3} & $+9.1$ \\
U-Net & 52.6 & \textbf{60.4} & $+7.8$ \\
DynUNet (4 levels) & 67.2 & \textbf{71.5} & $+4.4$ \\
V-Net & 45.3 & \textbf{49.2} & $+3.9$ \\
U-Net (wide) & 69.1 & \textbf{71.4} & $+2.4$ \\
DynUNet (5 levels) & 70.7 & \textbf{72.4} & $+1.7$ \\
Basic U-Net (wide) & 59.5 & \textbf{61.2} & $+1.7$ \\
Residual U-Net (deep) & 62.5 & \textbf{63.5} & $+1.0$ \\
U-Net (deep) & 65.6 & \textbf{66.4} & $+0.8$ \\
\midrule
Median / mean gain & & & $+3.9$ / $+4.9$ \\
\midrule
\multicolumn{4}{@{}l}{\textit{Original-label arm below 45 Dice}} \\
SegResNet (tiny) & 6.6 & \textbf{30.4} & $+23.8$ \\
SegResNet & 43.3 & \textbf{50.3} & $+7.0$ \\
UNETR (small) & 34.9 & \textbf{36.5} & $+1.6$ \\
U-Net (tiny) & 23.7 & \textbf{25.2} & $+1.5$ \\
\bottomrule
\end{tabular}
\end{table}

\section{Baseline Methods}\label{app:baselines}

Correction methods start from the predictions of the same upstream segmentation
model for each structure and return binary masks. \method receives segmentation
probabilities; methods that need a binary input use the corresponding thresholded
mask. On the
TotalSegmentator benchmark the upstream model is MedSAM; on the external
benchmarks it is an automatic segmentation model. No method sees test labels.
Hyperparameters are selected per class on the validation split by grid search
over the ranges below and then applied unchanged to the test cases, in the same
way as $t^*$ for \method. Morphological closing, the strongest single traditional
method on ToothFairy (Appendix~\ref{app:identity}), is also run on the held-out
WORD and AMOS22 inputs of Appendix~\ref{app:transfer} with the same per-class
sweep.

\paragraph{Learned methods.} All three are trained on segmentation predictions
paired with their reference annotations, with the same subject-level splits as
\method. The \emph{single-pass U-Net} \citep{ronneberger2015unet} uses \method's
backbone and text conditioning and receives the same thresholded prediction as
the other correction baselines, but maps it to its reference annotation in one
deterministic forward pass, without diffusion or partial noising. It tests
whether a single forward pass is enough. \emph{Post-DAE}
\citep{larrazabal2020postdae} is a denoising autoencoder that maps an implausible
mask to a plausible one; we reimplement it in 3D. \emph{SegRefiner}
\citep{wang2023segrefiner} adjusts a coarse mask with a discrete diffusion
process conditioned on the image, trained on the same pairs. It was designed for
2D images, so we apply it slice by slice, with the image.

\paragraph{Traditional methods: morphology and connected components.} These use
only the mask. Morphological operations use a ball-shaped structuring element
with radius swept over $\{1,2,3,4,5\}$ voxels. \emph{Hole filling}
\citep{soille2004morphology} fills all background components not connected to
the volume border and has no parameters. \emph{Morphological closing}
\citep{serra1983image} is a dilation followed by an erosion. \emph{Largest
component} \citep{he2009fastccl} keeps the largest 26-connected component, which
is only valid for structures that form a single component. \emph{Convex hull}
\citep{barber1996quickhull} replaces the mask by its 3D convex hull.

\paragraph{Traditional methods: statistical shape models.} Both are fitted on
the training split and applied by projection. Shapes are signed distance
functions on a fixed grid, rigidly aligned to a per-class mean. The number of
modes is swept over $\{5,10,20,50\}$ or chosen to explain $95\%$ of variance.
\emph{ASM} \citep{cootes1995asm} projects the shape onto the leading modes.
\emph{Robust ASM} \citep{rogers2002robustasm} uses a trimmed least-squares fit
that down-weights outlying boundary points, with trimming fraction swept over
$\{0.1,0.2,0.3\}$. Both produce a result for $41$ of the $90$ TotalSegmentator
anatomical classes.

\paragraph{Traditional methods: image-based.} These two methods, like SegRefiner,
read image intensity; the remaining correction baselines, including the
single-pass U-Net, and \method use segmentation predictions without the image. We include
them because a practitioner would try them first. \emph{GMM graph-cut}
\citep{rother2004grabcut} runs iterative graph-cut segmentation with Gaussian
mixture appearance models, five components per mixture and three iterations.
\emph{Geodesic active contour} \citep{caselles1997geodesic} evolves an edge-based
contour, with balloon force swept over $\{-1,0,1\}$ and iteration budget over
$\{25,50,100,200\}$.

\paragraph{Additional baselines on the external benchmarks.} On the external
benchmarks, the comparison extends the learned and traditional methods above
with opening, dilation, erosion, opening followed by closing, 2D hole filling,
removal of small components, Gaussian, median and signed-distance smoothing, a
2D convex hull, a mean-shape and a PCA signed-distance shape model, Chan--Vese
and random walker segmentation, and the identity control of
Appendix~\ref{app:identity}. Each is tuned per class on the validation split.
Figure~\ref{fig:organ_results} shows the two strongest for each structure.
Unless stated otherwise, baselines use scikit-image \citep{vanderwalt2014scikit},
SciPy \citep{virtanen2020scipy} and SimpleITK on the stated input
representations and identical hardware.

\section{Corpus-Scale Annotation and Release}\label{app:downstream}

We run the pipeline on $45{,}000$ CT scans assembled from the public datasets
catalogued by \citet{rokuss2026voxtell}. The pipeline has three stages. The same
predict-then-correct design has been used to assemble large annotated
collections with limited human review \citep{li2025scalemai}.

\paragraph{Stage 1: prediction by a suite of specialist models.} These scans
have no reference annotations, so MedSAM with bounding-box prompts from the
reference is not available. Instead we predict each structure with the published
segmentation model that covers it best. The suite comprises the TotalSegmentator
task models \citep{wasserthal2023totalsegmentator}, VISTA3D \citep{he2024vista3d},
the DAP-Atlas pipeline \citep{jaus2023dapatlas} and further published whole-body
and organ-specific models, together with MedSAM \citep{ma2024medsam} on the
ToothFairy structures, where reference boxes are available. Models were
assigned to structures by coverage of the structure, the modality and field of
view of the scan, and reported performance on that structure; where more than one
model covered a structure, the assigned model's prediction was used without
fusion. The full structure-to-model assignment table is released with the
annotations. Most scans cover only part of the body, so each contributes far
fewer classes than the vocabulary holds.

\paragraph{Stage 2: correction.} \method corrects each predicted probability map
with that class's description and its fixed $t^*$, with no retuning for the
corpus. The resulting corrected masks serve as training labels.

\begin{figure}[htbp]
\centering
\includegraphics[width=0.92\linewidth]{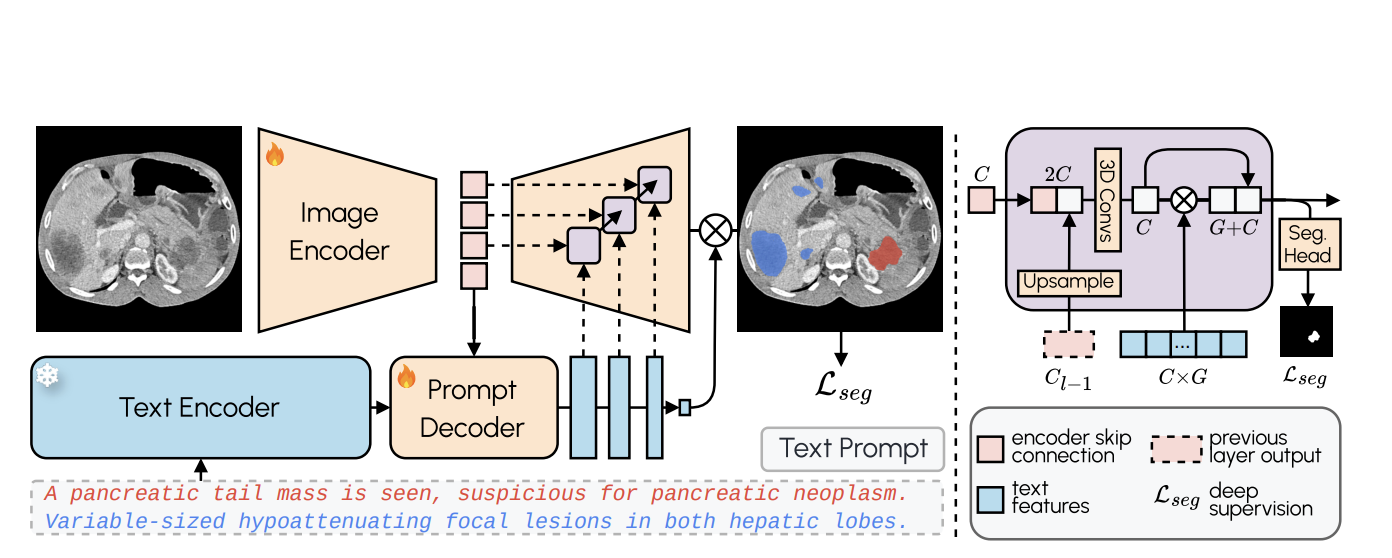}
\caption{\textbf{Architecture of \segmodel} \citep{rokuss2026voxtell}.
\textbf{Left:} a frozen text encoder embeds a text prompt naming the target
structure, and a trainable Prompt Decoder produces text features that modulate
the image decoder. \textbf{Right:} at each decoder level, upsampled features are
concatenated with the encoder skip connection and modulated by the text features
before the segmentation head, with deep supervision at every level. We use the
architecture unmodified and train it on \method-corrected training labels.}
\label{fig:voxtell}
\end{figure}

\paragraph{Stage 3: distillation and release.} \segmodel, a free-text
promptable 3D segmentation model with the architecture of
\citet{rokuss2026voxtell}, is trained on the corrected training labels and
applied across the corpus to produce the released annotations
(Figure~\ref{fig:voxtell}). Like \method, \segmodel names classes with text, so a
large vocabulary needs no equally large output layer. The released annotations
are therefore the output of \segmodel.

\paragraph{Compute.} \method takes $0.59$\,s per probability map on one AMD
MI250X GPU at \cropvol. The corpus pipeline took about two thousand GPU-hours on
AMD MI250X GPUs, and training the model, including the scaling and conditioning
experiments, a few hundred more.

\paragraph{Quality audit.} We drew a uniform random sample of $10{,}000$ of the
$45{,}000$ scans and split it into five disjoint blocks of $2{,}000$, one per
reader. For each scan, the reader saw the annotation with a written annotation
guideline for each structure present, and marked the scan as passing if every
structure followed its guideline. $98.5\%$ of scans passed, with a binomial
standard error below $0.2$ percentage points. Automated critics offer a
complementary route to flagging label errors in released collections
\citep{bassi2025label}.

\paragraph{Release.} We release the annotations for the $45{,}000$ scans,
together with the vocabulary, the description used for every class, the
per-class $t^*$ and the structure-to-model assignment table, so that the
annotations can be regenerated and audited independently. The source datasets
carry different licences, several of them non-commercial or share-alike, so
\dataset is released as an annotation layer for each source dataset under that
dataset's own terms.

\section{Anatomical Descriptions}\label{app:descriptions}

\begin{table}[!htbp]
\centering\small
\setlength{\tabcolsep}{5pt}
\caption{\textbf{Example anatomical descriptions}, one per anatomical group. The
last row is a held-out lesion class, described at inference only.}
\label{tab:description_examples}
\begin{tabularx}{\linewidth}{@{}llX@{}}
\toprule
Group & Anatomical class & Description \\
\midrule
Solid viscus & \texttt{pancreas} & pancreas, elongated lobulated gland, retroperitoneum behind the stomach from duodenal curve to splenic hilum \\
Hollow organ & \texttt{colon} & colon, long convoluted haustrated hollow tube, framing the peripheral abdomen from cecum to sigmoid \\
Gland & \texttt{adrenal\_gland\_left} & left adrenal gland, small thin comma-shaped gland with slender limbs, superior pole of the left kidney medial to the spleen \\
Vessel & \texttt{aorta} & aorta, thick-walled large arterial tube arching then descending, from the left ventricle through the mediastinum to the iliac bifurcation \\
Muscle & \texttt{autochthon\_left} & left autochthon, long paraspinal erector spinae muscle mass, alongside the left vertebral column from sacrum to skull base \\
Bone & \texttt{clavicula\_left} & left clavicle, slender S-shaped long bone, horizontal at the left base of the neck from sternum to scapula \\
Neuro / ocular & \texttt{chiasm} & optic chiasm, small flat X-shaped neural crossing, skull base above the pituitary gland \\
Lesion (held out) & \texttt{kidney\_cyst\_left} & left kidney cyst, rounded fluid-filled lesion, within or on the surface of the left kidney \\
\bottomrule
\end{tabularx}
\end{table}

Anatomical descriptions identify each class and characterize its shape and
location; their level of detail can vary with the anatomy being described. For
the experiments reported here, descriptions were drafted by a large language
model from standard anatomy references and reviewed by the authors before
training. The same process was used for lesion descriptions at inference.
Table~\ref{tab:description_examples} gives one example per anatomical group; the
full set is released with the code, since the vocabulary is large enough that a
printed table would be neither readable nor a reliable record of what was used.

\section{Per-Class Comparison Across All Methods}\label{app:pergroup}

Table~\ref{tab:allstructures} compares all methods on the $90$ TotalSegmentator
anatomical classes, in the format of the Touchstone benchmark
\citep{bassi2024touchstone}: classes across the top, methods down the side, read
in blocks of six. Every method starts from the MedSAM predictions.
Table~\ref{tab:headline} in the main
text shows the six classes with the largest \method gains.

{\scriptsize
\setlength{\tabcolsep}{2pt}
\renewcommand{\arraystretch}{1.02}
\begin{longtable}{@{}l*{6}{c}@{}}
\caption{\textbf{Dice for all methods on the $90$ TotalSegmentator anatomical
classes.} Read in blocks of six; each column is one anatomical class. Every
method starts from the MedSAM predictions. Best
per column in \best{bold blue}, second best in \second{blue}. `---' means the
method produces no result (the two ASM variants cover $41$ of the $90$ classes).
$^\dagger$ reads image intensity. \method is the only method whose Dice is never
below the input.}
\label{tab:allstructures}\\
\toprule
\endfirsthead
\multicolumn{7}{@{}l}{\itshape Table~\ref{tab:allstructures} continued from previous page}\\
\toprule
\endhead
\bottomrule
\multicolumn{7}{@{}r}{\itshape continued on next page}\\
\endfoot
\bottomrule
\endlastfoot

Method & \textbf{Adrenal L} & \textbf{Adrenal R} & \textbf{Aorta} & \textbf{Atr. app. L} & \textbf{Autoch. L} & \textbf{Autoch. R} \\
\midrule
MedSAM (input) & 78.6 & 65.6 & 85.1 & 74.6 & 79.1 & 79.8 \\
\midrule
Single-pass U-Net & 78.6 & 65.6 & 84.5 & 74.6 & 78.8 & 79.5 \\
SegRefiner$^\dagger$ & 72.3 & 64.6 & 74.9 & 70.2 & 75.1 & 74.9 \\
Post-DAE & 49.0 & 39.0 & 53.6 & 41.7 & 62.4 & 62.7 \\
Hole filling & 78.6 & 65.6 & 85.1 & 74.6 & 79.1 & 79.8 \\
Morph. closing & \second{79.3} & \second{66.6} & 83.8 & 76.0 & 81.8 & \second{82.2} \\
Geodesic$^\dagger$ & 17.3 & 1.4 & 67.2 & 18.2 & 74.6 & 74.1 \\
GMM graph-cut$^\dagger$ & 74.3 & 57.9 & 55.2 & 61.5 & 66.8 & 66.2 \\
Convex hull & 77.4 & \second{66.6} & 81.3 & \second{76.2} & \second{81.9} & 82.0 \\
Largest comp. & 78.1 & 64.4 & \second{85.3} & 74.5 & 79.1 & 79.7 \\
ASM & --- & --- & --- & --- & 21.3 & 18.7 \\
Robust ASM & --- & --- & --- & --- & 20.7 & 18.3 \\
\midrule
\textbf{\method} & \best{80.6} & \best{70.7} & \best{85.5} & \best{79.2} & \best{85.7} & \best{85.4} \\
\midrule
Method & \textbf{Brachioc. trunk} & \textbf{Brachioc. vein L} & \textbf{Brachioc. vein R} & \textbf{Brain} & \textbf{Clavicle L} & \textbf{Clavicle R} \\
\midrule
MedSAM (input) & 85.1 & 79.5 & 85.4 & 85.6 & 83.8 & 82.8 \\
\midrule
Single-pass U-Net & 85.1 & 79.5 & 85.4 & \second{85.6} & 83.6 & 82.7 \\
SegRefiner$^\dagger$ & 77.8 & 75.5 & 77.5 & 80.8 & 74.2 & 74.0 \\
Post-DAE & 53.3 & 54.7 & 56.3 & 73.0 & 57.5 & 62.3 \\
Hole filling & 85.1 & 79.5 & 85.4 & \second{85.6} & \second{83.8} & \second{82.9} \\
Morph. closing & \best{86.7} & \second{81.0} & \second{86.8} & 84.9 & 82.9 & 81.5 \\
Geodesic$^\dagger$ & 33.9 & 42.0 & 27.6 & 55.7 & 66.3 & 64.0 \\
GMM graph-cut$^\dagger$ & 56.2 & 41.3 & 56.8 & 67.8 & 35.1 & 31.0 \\
Convex hull & 85.4 & 79.5 & 86.1 & 83.6 & 82.2 & 81.5 \\
Largest comp. & 85.1 & 79.5 & 85.4 & \second{85.6} & 83.7 & 82.8 \\
ASM & --- & --- & --- & --- & 10.8 & 12.2 \\
Robust ASM & --- & --- & --- & --- & 10.8 & 12.2 \\
\midrule
\textbf{\method} & \second{86.6} & \best{82.6} & \best{87.6} & \best{86.4} & \best{84.7} & \best{84.4} \\
\midrule
Method & \textbf{Colon} & \textbf{C. carotid L} & \textbf{C. carotid R} & \textbf{Costal c.} & \textbf{Duodenum} & \textbf{Esophagus} \\
\midrule
MedSAM (input) & 46.6 & 75.4 & 74.1 & 29.6 & 65.5 & 78.5 \\
\midrule
Single-pass U-Net & 46.4 & 75.4 & 74.2 & 29.0 & \second{65.5} & 78.0 \\
SegRefiner$^\dagger$ & 44.6 & 67.2 & 68.3 & 28.8 & 62.2 & 73.2 \\
Post-DAE & 33.7 & 51.2 & 50.4 & 17.6 & 56.4 & 58.0 \\
Hole filling & \second{46.6} & 75.4 & 74.1 & \second{29.6} & \second{65.5} & 78.5 \\
Morph. closing & 46.2 & \second{77.8} & \second{76.1} & 28.6 & 62.9 & \second{80.2} \\
Geodesic$^\dagger$ & 42.0 & 13.6 & 6.9 & 24.6 & 46.7 & 34.3 \\
GMM graph-cut$^\dagger$ & 42.8 & 56.9 & 54.6 & 20.1 & 61.3 & 72.2 \\
Convex hull & 39.5 & 77.1 & 75.4 & 21.9 & 61.6 & \second{80.2} \\
Largest comp. & 45.0 & 71.4 & 71.5 & 27.8 & 64.3 & 74.1 \\
ASM & 12.4 & 6.6 & 5.9 & --- & --- & 6.6 \\
Robust ASM & 12.4 & 6.6 & 5.9 & --- & --- & 6.6 \\
\midrule
\textbf{\method} & \best{47.1} & \best{79.8} & \best{78.3} & \best{30.3} & \best{67.4} & \best{83.3} \\
\midrule
Method & \textbf{Femur L} & \textbf{Femur R} & \textbf{Gallbladder} & \textbf{Glut. max. L} & \textbf{Glut. max. R} & \textbf{Glut. med. L} \\
\midrule
MedSAM (input) & 90.7 & 91.0 & 90.9 & 89.9 & 89.5 & 81.1 \\
\midrule
Single-pass U-Net & 90.7 & 90.9 & \second{90.9} & 89.9 & \second{89.6} & \second{81.2} \\
SegRefiner$^\dagger$ & 83.6 & 83.3 & 84.9 & 82.4 & 82.9 & 78.7 \\
Post-DAE & 67.5 & 65.9 & 62.1 & 65.7 & 64.9 & 71.2 \\
Hole filling & \second{91.1} & \second{91.4} & \second{90.9} & 89.9 & 89.5 & 81.1 \\
Morph. closing & 90.1 & 89.4 & 88.4 & \second{90.0} & 89.3 & 81.1 \\
Geodesic$^\dagger$ & 82.1 & 83.8 & 53.3 & 83.8 & 83.3 & 69.8 \\
GMM graph-cut$^\dagger$ & 70.7 & 70.5 & 85.4 & 70.0 & 72.3 & 60.6 \\
Convex hull & 90.3 & 90.3 & 90.8 & 88.3 & 87.5 & 79.8 \\
Largest comp. & 90.7 & 91.0 & \second{90.9} & 89.9 & 89.5 & 81.1 \\
ASM & 37.2 & 36.5 & --- & 29.5 & 25.8 & 23.7 \\
Robust ASM & 37.2 & 36.5 & --- & 29.5 & 25.8 & 23.7 \\
\midrule
\textbf{\method} & \best{91.5} & \best{92.0} & \best{92.0} & \best{90.9} & \best{90.4} & \best{83.1} \\
\midrule
Method & \textbf{Glut. med. R} & \textbf{Glut. min. L} & \textbf{Glut. min. R} & \textbf{Heart} & \textbf{Hip L} & \textbf{Hip R} \\
\midrule
MedSAM (input) & 80.8 & 63.8 & 64.8 & 87.2 & 62.5 & 59.9 \\
\midrule
Single-pass U-Net & 80.8 & 63.7 & 64.9 & 87.0 & 62.3 & 59.6 \\
SegRefiner$^\dagger$ & 78.2 & 64.2 & 64.6 & 79.1 & 58.7 & 57.2 \\
Post-DAE & 68.8 & 51.4 & 56.3 & 65.7 & 52.9 & 51.9 \\
Hole filling & 80.8 & 63.8 & 64.8 & \second{87.2} & 62.5 & 59.9 \\
Morph. closing & 80.6 & \second{66.5} & \second{65.5} & 85.6 & 61.3 & \second{60.2} \\
Geodesic$^\dagger$ & 70.0 & 19.9 & 25.4 & 81.7 & \best{64.4} & \best{63.3} \\
GMM graph-cut$^\dagger$ & 63.3 & 64.0 & 65.3 & 62.2 & 49.4 & 45.9 \\
Convex hull & 79.3 & 64.6 & 63.8 & 85.7 & 59.6 & 57.2 \\
Largest comp. & \second{80.9} & 63.8 & 64.8 & \second{87.2} & 62.5 & 59.9 \\
ASM & 17.7 & 11.1 & 20.4 & --- & 15.5 & 22.5 \\
Robust ASM & 17.7 & 11.1 & 20.4 & --- & 15.5 & 22.5 \\
\midrule
\textbf{\method} & \best{82.1} & \best{68.8} & \best{67.2} & \best{87.7} & \second{63.0} & 60.1 \\
\midrule
Method & \textbf{Humerus L} & \textbf{Humerus R} & \textbf{Iliac a. L} & \textbf{Iliac a. R} & \textbf{Iliac v. L} & \textbf{Iliac v. R} \\
\midrule
MedSAM (input) & 88.9 & 87.0 & 80.2 & 78.6 & 82.7 & 84.7 \\
\midrule
Single-pass U-Net & 88.9 & 87.0 & 79.0 & 77.2 & 82.1 & 84.1 \\
SegRefiner$^\dagger$ & 79.5 & 79.5 & 70.3 & 67.6 & 76.9 & 78.1 \\
Post-DAE & 53.2 & 50.6 & 62.4 & 64.1 & 68.1 & 61.5 \\
Hole filling & 89.2 & 87.2 & 80.2 & 78.6 & 82.7 & 84.7 \\
Morph. closing & 83.8 & 76.8 & 80.1 & \second{79.5} & \second{84.5} & \second{86.1} \\
Geodesic$^\dagger$ & 65.7 & 68.5 & 22.8 & 21.9 & 36.4 & 29.1 \\
GMM graph-cut$^\dagger$ & 79.9 & 81.8 & 51.5 & 49.4 & 57.0 & 52.9 \\
Convex hull & \second{90.0} & \second{87.9} & \second{80.5} & 79.2 & 83.4 & 85.3 \\
Largest comp. & 88.9 & 87.0 & 76.4 & 71.9 & 81.8 & 84.0 \\
ASM & --- & --- & 4.0 & 3.5 & 5.6 & 6.1 \\
Robust ASM & --- & --- & 4.0 & 3.5 & 5.6 & 6.1 \\
\midrule
\textbf{\method} & \best{91.0} & \best{89.0} & \best{81.6} & \best{80.8} & \best{85.3} & \best{87.0} \\
\midrule
Method & \textbf{Iliopsoas L} & \textbf{Iliopsoas R} & \textbf{IVC} & \textbf{Kidney L} & \textbf{Kidney R} & \textbf{Liver} \\
\midrule
MedSAM (input) & 80.6 & 80.7 & 88.7 & 94.4 & 96.2 & 95.8 \\
\midrule
Single-pass U-Net & 80.4 & 80.5 & 88.2 & 94.3 & 96.1 & 95.3 \\
SegRefiner$^\dagger$ & 76.3 & 75.8 & 81.3 & 85.6 & 86.6 & 86.7 \\
Post-DAE & 69.7 & 70.0 & 65.3 & 70.4 & 69.6 & 77.2 \\
Hole filling & 80.6 & 80.7 & 88.7 & \second{94.4} & 96.1 & \second{95.8} \\
Morph. closing & \second{81.5} & \second{81.0} & \second{89.9} & 93.4 & 94.3 & 94.8 \\
Geodesic$^\dagger$ & 58.7 & 57.9 & 54.1 & 84.0 & 84.1 & 88.6 \\
GMM graph-cut$^\dagger$ & 66.5 & 75.5 & 73.2 & 74.3 & 78.8 & 63.7 \\
Convex hull & 79.7 & 80.0 & 89.1 & 90.1 & 92.1 & 90.3 \\
Largest comp. & 79.0 & 79.5 & 87.7 & \second{94.4} & \second{96.2} & \best{95.9} \\
ASM & 21.9 & 15.0 & --- & --- & --- & --- \\
Robust ASM & 21.9 & 15.0 & --- & --- & --- & --- \\
\midrule
\textbf{\method} & \best{82.4} & \best{82.6} & \best{90.3} & \best{94.7} & \best{96.4} & \second{95.8} \\
\midrule
Method & \textbf{Lung low. L} & \textbf{Lung low. R} & \textbf{Lung mid. R} & \textbf{Lung up. L} & \textbf{Lung up. R} & \textbf{Pancreas} \\
\midrule
MedSAM (input) & 72.9 & 68.8 & 77.5 & 78.4 & 79.8 & 80.1 \\
\midrule
Single-pass U-Net & 73.0 & 68.9 & \second{77.6} & 78.3 & 79.8 & 80.0 \\
SegRefiner$^\dagger$ & 69.4 & 63.7 & 71.0 & 71.4 & 73.1 & 73.0 \\
Post-DAE & 58.0 & 51.8 & 58.9 & 60.2 & 64.4 & 63.2 \\
Hole filling & 72.9 & 68.8 & 77.5 & 78.4 & 79.8 & 80.1 \\
Morph. closing & 72.9 & 67.6 & 72.5 & 76.4 & 75.6 & 77.2 \\
Geodesic$^\dagger$ & 73.6 & \best{74.3} & 70.1 & 75.5 & 71.4 & 59.1 \\
GMM graph-cut$^\dagger$ & \best{75.3} & \second{71.7} & 76.6 & \best{81.1} & \best{82.8} & 78.7 \\
Convex hull & 72.6 & 70.6 & 77.2 & 77.8 & 79.3 & 69.3 \\
Largest comp. & 72.9 & 68.8 & 77.5 & 78.3 & 79.8 & \second{80.2} \\
ASM & --- & --- & --- & 18.8 & 18.0 & --- \\
Robust ASM & --- & --- & --- & 18.8 & 18.0 & --- \\
\midrule
\textbf{\method} & \second{74.2} & 70.2 & \best{78.6} & \second{79.1} & \second{80.7} & \best{80.6} \\
\midrule
Method & \textbf{Portal v.} & \textbf{Prostate} & \textbf{Pulm. v.} & \textbf{Rib L 1} & \textbf{Rib L 10} & \textbf{Rib L 11} \\
\midrule
MedSAM (input) & 35.6 & 82.2 & 37.8 & 74.3 & 64.7 & 65.0 \\
\midrule
Single-pass U-Net & 35.3 & 82.2 & 37.7 & 74.4 & 63.3 & 64.6 \\
SegRefiner$^\dagger$ & 33.0 & 79.8 & 37.0 & 68.0 & 58.3 & 59.3 \\
Post-DAE & 28.7 & 57.4 & 22.2 & 60.0 & 56.2 & 59.0 \\
Hole filling & 35.6 & 82.2 & \second{37.8} & 74.3 & 64.7 & 65.0 \\
Morph. closing & 35.4 & \best{85.7} & 30.9 & \best{76.1} & \second{65.8} & \second{66.2} \\
Geodesic$^\dagger$ & 22.5 & 66.0 & 19.8 & 21.9 & 31.8 & 21.5 \\
GMM graph-cut$^\dagger$ & \best{37.8} & 79.8 & 28.5 & 33.9 & 20.8 & 19.3 \\
Convex hull & 28.9 & 84.0 & 35.2 & 73.3 & 63.6 & 64.1 \\
Largest comp. & 32.3 & 82.2 & 36.8 & 73.2 & 62.6 & 59.7 \\
ASM & --- & --- & --- & 7.9 & --- & --- \\
Robust ASM & --- & --- & --- & 7.9 & --- & --- \\
\midrule
\textbf{\method} & \second{35.8} & \second{85.5} & \best{38.6} & \second{75.8} & \best{68.3} & \best{68.8} \\
\midrule
Method & \textbf{Rib L 12} & \textbf{Rib L 2} & \textbf{Rib L 3} & \textbf{Rib L 4} & \textbf{Rib L 5} & \textbf{Rib L 6} \\
\midrule
MedSAM (input) & 63.6 & 71.7 & 63.4 & 64.2 & 65.9 & 67.9 \\
\midrule
Single-pass U-Net & 63.6 & 71.3 & 62.8 & 63.2 & 64.9 & 67.3 \\
SegRefiner$^\dagger$ & 55.2 & 66.8 & 57.7 & 59.0 & 59.4 & 61.5 \\
Post-DAE & 50.8 & 53.6 & 53.5 & 52.9 & 53.0 & 48.5 \\
Hole filling & 63.6 & 71.7 & \second{63.4} & \second{64.2} & \second{65.9} & 67.9 \\
Morph. closing & \second{65.0} & \second{72.2} & 62.9 & 59.8 & 65.6 & \second{68.4} \\
Geodesic$^\dagger$ & 9.3 & 18.6 & 14.7 & 18.3 & 27.7 & 37.0 \\
GMM graph-cut$^\dagger$ & 28.7 & 27.4 & 22.8 & 23.2 & 22.8 & 26.8 \\
Convex hull & 62.5 & 69.8 & 61.2 & 62.3 & 64.0 & 67.2 \\
Largest comp. & 60.7 & 66.3 & 59.5 & 63.8 & 63.9 & 66.4 \\
ASM & --- & 6.3 & --- & --- & --- & --- \\
Robust ASM & --- & 6.3 & --- & --- & --- & --- \\
\midrule
\textbf{\method} & \best{66.5} & \best{73.8} & \best{65.3} & \best{66.5} & \best{67.7} & \best{70.9} \\
\midrule
Method & \textbf{Rib L 7} & \textbf{Rib L 8} & \textbf{Rib L 9} & \textbf{Rib R 1} & \textbf{Rib R 10} & \textbf{Rib R 11} \\
\midrule
MedSAM (input) & 66.2 & 62.2 & 62.5 & 76.5 & 72.3 & 70.6 \\
\midrule
Single-pass U-Net & 65.3 & 61.6 & 61.3 & 76.5 & 70.7 & 69.8 \\
SegRefiner$^\dagger$ & 60.6 & 56.7 & 57.5 & 70.5 & 64.4 & 63.4 \\
Post-DAE & 47.8 & 45.9 & 51.2 & 59.3 & 63.7 & 63.6 \\
Hole filling & 66.2 & \second{62.2} & 62.5 & 76.5 & 72.3 & 70.6 \\
Morph. closing & \second{67.9} & 61.3 & \second{64.3} & \best{78.1} & \second{73.0} & \second{71.9} \\
Geodesic$^\dagger$ & 34.1 & 30.3 & 33.9 & 24.8 & 27.1 & 23.1 \\
GMM graph-cut$^\dagger$ & 25.9 & 29.5 & 22.7 & 35.9 & 19.5 & 18.2 \\
Convex hull & 65.3 & 61.6 & 61.1 & 75.9 & 70.8 & 69.2 \\
Largest comp. & 63.9 & 60.2 & 59.4 & 74.8 & 71.8 & 69.3 \\
ASM & --- & --- & --- & 6.8 & --- & --- \\
Robust ASM & --- & --- & --- & 6.8 & --- & --- \\
\midrule
\textbf{\method} & \best{70.1} & \best{65.4} & \best{65.5} & \second{77.6} & \best{75.1} & \best{73.4} \\
\midrule
Method & \textbf{Rib R 12} & \textbf{Rib R 2} & \textbf{Rib R 3} & \textbf{Rib R 4} & \textbf{Rib R 5} & \textbf{Rib R 6} \\
\midrule
MedSAM (input) & 68.7 & 73.8 & 62.6 & 66.2 & 67.0 & 70.9 \\
\midrule
Single-pass U-Net & 68.6 & 73.6 & 61.4 & 65.5 & 66.1 & 70.3 \\
SegRefiner$^\dagger$ & 58.3 & 70.1 & 57.8 & 60.1 & 60.6 & 63.2 \\
Post-DAE & 55.7 & 60.1 & 50.0 & 55.6 & 53.5 & 49.2 \\
Hole filling & 68.7 & 73.8 & \second{62.6} & \second{66.2} & 67.0 & \second{70.9} \\
Morph. closing & \second{70.9} & \best{75.9} & 62.0 & 63.6 & \second{67.8} & 70.2 \\
Geodesic$^\dagger$ & 9.8 & 18.2 & 14.0 & 21.2 & 29.6 & 34.3 \\
GMM graph-cut$^\dagger$ & 29.1 & 23.5 & 22.7 & 22.7 & 19.5 & 28.4 \\
Convex hull & 68.2 & 72.1 & 59.6 & 64.0 & 65.2 & 70.0 \\
Largest comp. & 63.5 & 70.4 & 51.9 & 62.5 & 66.8 & 69.3 \\
ASM & --- & 7.0 & --- & --- & --- & --- \\
Robust ASM & --- & 7.0 & --- & --- & --- & --- \\
\midrule
\textbf{\method} & \best{72.4} & \second{75.7} & \best{64.9} & \best{68.4} & \best{69.2} & \best{72.9} \\
\midrule
Method & \textbf{Rib R 7} & \textbf{Rib R 8} & \textbf{Rib R 9} & \textbf{Sacrum} & \textbf{Scapula L} & \textbf{Scapula R} \\
\midrule
MedSAM (input) & 70.8 & 71.3 & 70.7 & 64.3 & 51.0 & 48.6 \\
\midrule
Single-pass U-Net & 69.9 & 70.3 & 69.7 & 64.4 & \second{51.0} & 48.4 \\
SegRefiner$^\dagger$ & 64.5 & 62.9 & 63.4 & 65.0 & 48.7 & 45.9 \\
Post-DAE & 50.9 & 54.2 & 59.4 & 56.8 & 37.5 & 34.6 \\
Hole filling & 70.8 & 71.3 & 70.7 & 64.4 & \second{51.0} & \second{48.6} \\
Morph. closing & \second{71.3} & \second{71.9} & \second{71.0} & \second{65.3} & 48.3 & 44.8 \\
Geodesic$^\dagger$ & 33.6 & 29.8 & 33.2 & 59.9 & 37.9 & 36.9 \\
GMM graph-cut$^\dagger$ & 28.1 & 25.9 & 22.8 & 48.1 & 44.5 & 43.4 \\
Convex hull & 69.5 & 69.5 & 69.0 & 61.6 & 49.7 & 47.9 \\
Largest comp. & 70.6 & 68.0 & 70.2 & 64.3 & 50.9 & 48.5 \\
ASM & --- & --- & --- & 17.3 & 12.1 & 9.4 \\
Robust ASM & --- & --- & --- & 17.3 & 12.1 & 9.4 \\
\midrule
\textbf{\method} & \best{73.0} & \best{74.5} & \best{73.0} & \best{65.8} & \best{51.1} & \best{49.2} \\
\midrule
Method & \textbf{Skull} & \textbf{Small bowel} & \textbf{Spinal cord} & \textbf{Spleen} & \textbf{Sternum} & \textbf{Stomach} \\
\midrule
MedSAM (input) & 60.5 & 63.6 & 77.5 & 96.4 & 78.2 & 90.8 \\
\midrule
Single-pass U-Net & 60.3 & 63.5 & 76.9 & 96.1 & 78.0 & 90.5 \\
SegRefiner$^\dagger$ & 50.9 & 60.9 & 73.9 & 86.8 & 72.8 & 81.5 \\
Post-DAE & 31.0 & 51.2 & 61.8 & 74.4 & 52.5 & 68.2 \\
Hole filling & 60.5 & 63.6 & 77.5 & \second{96.4} & 78.2 & \second{90.8} \\
Morph. closing & 49.5 & \second{64.1} & \second{80.0} & 95.1 & \second{80.2} & 90.6 \\
Geodesic$^\dagger$ & 45.3 & 56.6 & 63.6 & 84.0 & 57.7 & 77.7 \\
GMM graph-cut$^\dagger$ & \second{62.7} & 62.0 & 77.4 & 64.6 & 58.5 & 88.0 \\
Convex hull & 60.2 & 58.8 & 78.5 & 91.6 & 78.7 & 87.9 \\
Largest comp. & 60.5 & 63.5 & 72.6 & \second{96.4} & 78.0 & \second{90.8} \\
ASM & 16.6 & 12.5 & 8.2 & --- & --- & --- \\
Robust ASM & 16.6 & 12.5 & 6.3 & --- & --- & --- \\
\midrule
\textbf{\method} & \best{63.0} & \best{64.8} & \best{83.1} & \best{96.5} & \best{82.4} & \best{92.0} \\
\midrule
Method & \textbf{Subclav. L} & \textbf{Subclav. R} & \textbf{SVC} & \textbf{Thyroid} & \textbf{Trachea} & \textbf{Bladder} \\
\midrule
MedSAM (input) & 54.1 & 57.4 & 86.6 & 51.5 & 78.7 & 90.1 \\
\midrule
Single-pass U-Net & 54.1 & 57.3 & 86.6 & 51.5 & 78.8 & 90.0 \\
SegRefiner$^\dagger$ & 50.4 & 55.7 & 80.8 & 52.3 & 73.1 & 85.0 \\
Post-DAE & 36.6 & 37.7 & 58.6 & 35.4 & 65.6 & 71.1 \\
Hole filling & 54.1 & 57.4 & 86.6 & 51.5 & 78.7 & 90.2 \\
Morph. closing & \second{54.3} & \second{58.8} & 81.9 & \second{53.0} & \second{80.2} & 89.7 \\
Geodesic$^\dagger$ & 10.9 & 12.4 & 57.9 & 22.0 & 63.1 & 79.8 \\
GMM graph-cut$^\dagger$ & 48.1 & 56.9 & 50.3 & 47.5 & 67.7 & 83.6 \\
Convex hull & 51.0 & 53.9 & \second{88.0} & 46.6 & 79.0 & \second{91.2} \\
Largest comp. & 48.6 & 56.6 & 86.6 & 51.1 & 78.7 & 90.2 \\
ASM & 5.1 & 8.1 & --- & 10.6 & 12.8 & 29.6 \\
Robust ASM & 5.1 & 8.1 & --- & 10.6 & 12.8 & 29.6 \\
\midrule
\textbf{\method} & \best{55.5} & \best{58.9} & \best{90.0} & \best{53.5} & \best{80.7} & \best{92.6} \\

\end{longtable}
}

\end{document}